\documentclass[sigconf]{acmart}
\usepackage{booktabs} % For formal tables
\usepackage{amsmath,amssymb,amsfonts}
\usepackage{algorithmic}
\usepackage{graphicx}
\usepackage{textcomp}
\usepackage{xcolor}
\usepackage{bm}
\usepackage{algorithmic, algorithm}
\usepackage{subfig}
\usepackage{pgfplots}
\usepackage{resizegather}
\usepackage{multirow}
\newtheorem{definition}{Definition}

\newcommand{\sysname}{CSIM}
\DeclareMathOperator*{\argmax}{argmax}
\DeclareMathOperator*{\argmin}{argmin}
\def\BibTeX{{\rm B\kern-.05em{\sc i\kern-.025em b}\kern-.08em
    T\kern-.1667em\lower.7ex\hbox{E}\kern-.125emX}}
    
\setcopyright{rightsretained}
\acmDOI{10.475/123_4}

\acmISBN{123-000--567/08/06}

\acmConference[KDD'18 Deep Learning Day]{ACM KDD'18 Deep Learning Day}{August 2018}{London, UK}
\acmYear{2018}
\copyrightyear{2018}
\acmArticle{4}
\acmPrice{15.00}

\editor{Jennifer B. Sartor}
\editor{Theo D'Hondt}
\editor{Wolfgang De Meuter}

\begin{document}
% \title{\sysname{}: Convolutional Open-World Multi-Task Image Stream Classifier with Intrinsic Similarity Metrics}
\title[Adaptive Image Stream Classification with Conv-OWC]{Adaptive Image Stream Classification via Convolutional Neural Network with Intrinsic Similarity Metrics}
% \titlenote{Produces the permission block, and
%   copyright information}
% \subtitle{Extended Abstract}
% \subtitlenote{The full version of the author's guide is available as
%   \texttt{acmart.pdf} document}

% \author{Yang Gao; Swarup Chandra; Zhuoyi Wang; Latifur Khan}
% % \authornote{Dr.~Trovato insisted his name be first.}
% % \orcid{1234-5678-9012}
% \affiliation{%
%   \institution{University of Texas at Dallas, USA \\ \{yxg122530; src093020; zxw151030; lkhan\}@utdallas.edu}
% %   \streetaddress{P.O. Box 1212}
% %   \city{Dublin}
% %   \state{Ohio}
% %   \postcode{43017-6221}
% }
% % \email{{yxg122530; src093020; zxw151030; lkhan}@utdallas.edu}

\author{Yang Gao}
% \authornote{This author is the one did all the really hard work.}
\affiliation{%
  \institution{University of Texas at Dallas}
%   \streetaddress{P.O. Box 1212}
  \city{Dallas}
  \state{Texas}
  \country{USA}
%   \postcode{43017-6221}
}
\email{yxg122530@utdallas.edu}

\author{Swarup Chandra}
% \authornote{The secretary disavows any knowledge of this author's actions.}
\affiliation{%
  \institution{University of Texas at Dallas}
%   \streetaddress{P.O. Box 1212}
  \city{Dallas}
  \state{Texas}
  \country{USA}
%   \postcode{43017-6221}
}
\email{swarup.chandra@utdallas.edu}

\author{Zhuoyi Wang}
\affiliation{%
	\institution{University of Texas at Dallas}
    \city{Dallas}
    \state{Texas}
    \country{USA}
}
\email{Zhuoyi.Wang1@utdallas.edu}

\author{Latifur Khan}
% \authornote{This author is the
%   one who did all the really hard work.}
\affiliation{%
  \institution{University of Texas at Dallas}
  \city{Dallas}
  \state{Texas}
  \country{USA}
%   \streetaddress{1 Th{\o}rv{\"a}ld Circle}
%   \city{Hekla}
%   \country{Iceland}
}
\email{lkhan@utdallas.edu}

% \author{Valerie B\'eranger}
% \affiliation{%
%   \institution{Inria Paris-Rocquencourt}
%   \city{Rocquencourt}
%   \country{France}
% }
% \author{Aparna Patel}
% \affiliation{%
%  \institution{Rajiv Gandhi University}
%  \streetaddress{Rono-Hills}
%  \city{Doimukh}
%  \state{Arunachal Pradesh}
%  \country{India}}
% \author{Huifen Chan}
% \affiliation{%
%   \institution{Tsinghua University}
%   \streetaddress{30 Shuangqing Rd}
%   \city{Haidian Qu}
%   \state{Beijing Shi}
%   \country{China}
% }

% \author{Charles Palmer}
% \affiliation{%
%   \institution{Palmer Research Laboratories}
%   \streetaddress{8600 Datapoint Drive}
%   \city{San Antonio}
%   \state{Texas}
%   \postcode{78229}}
% \email{cpalmer@prl.com}

% \author{John Smith}
% \affiliation{\institution{The Th{\o}rv{\"a}ld Group}}
% \email{jsmith@affiliation.org}

% \author{Julius P.~Kumquat}
% \affiliation{\institution{The Kumquat Consortium}}
% \email{jpkumquat@consortium.net}

% The default list of authors is too long for headers.
% \renewcommand{\shortauthors}{B. Trovato et al.}

\begin{abstract}
When performing data classification over a stream of continuously occurring instances, a key challenge is to develop an open-world classifier that anticipates instances from an unknown class. Studies addressing this problem, typically called novel class detection, have considered classification methods that reactively adapt to such changes along the stream. Importantly, they rely on the property of cohesion and separation among instances in feature space. Instances belonging to the same class are assumed to be closer to each other (cohesion) than those belonging to different classes (separation). Unfortunately, this assumption may not have large support when dealing with high dimensional data such as images. In this paper, we address this key challenge by proposing a semi-supervised multi-task learning framework called \sysname{} which aims to intrinsically search for a latent space suitable for detecting labels of instances from both known and unknown classes. Particularly, we utilize a convolution neural network layer that aids in the learning of a latent feature space suitable for novel class detection. We empirically measure the performance of \sysname{} over multiple real-world image datasets and demonstrate its superiority by comparing its performance with existing semi-supervised methods. 
\end{abstract}



\maketitle

\section{Introduction}
\label{Sec:introduction}
%   Primary contributions of our work are as follows: (a)
% The recent proliferation of Internet technology in daily life has generated an enormous amount of data streams from various sources such as social networks, online business, sensors, and online news-feed. Therefore, having a robust and high-performance technique for classifying data streams is becoming increasingly important.  However, classification of data streams is challenging due to its inherent properties such as infinite length, concept drift, concept evolution, and limited labeled data.
% The problem of infinite length is typically addressed by observing the data stream either in a fixed-size chunk (e.g., \cite{parker2015}) or as a stream with individual data instances continuously arriving (e.g., \cite{BeringerH07}). 
% Furthermore, a dynamic sliding window (e.g. \cite{bifet2007}) is used to track changes in classifier performance for drift detection and classifier adaptation. This occurs due to the changes in data distribution over time.

A stream of data typically results from applications such as social networks, online business transactions, news-feeds etc.
Recent studies have attempted to address the infinite length challenge by employing a fixed-size sliding window to perform analytics\cite{BeringerH07,bifet2007, parker2015}. In particular, the data sources are assumed to be non-stationary whose data distribution changes over time. This property directly affects a trained classifier. Therefore, a reactive mechanism is typically used where a change is first detected and then appropriate actions to adapt the classifier are considered.
 
\begin{figure}%
    \centering
    \subfloat[\label{fig:digit_similarity}]{{\includegraphics[width=0.7\columnwidth]{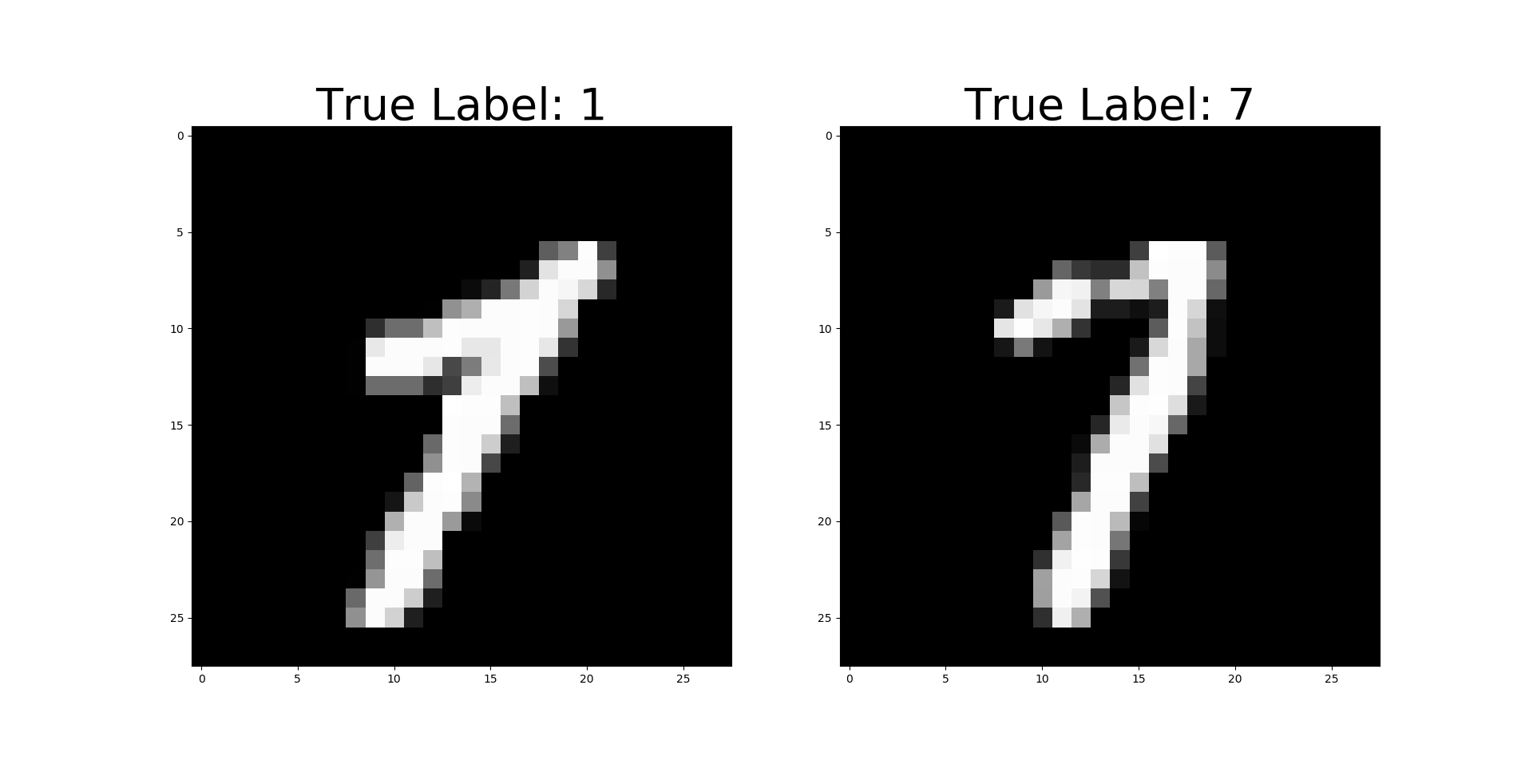} }}%
%     \qquad
%     \subfloat[\label{fig:metric_concept}]{{\includegraphics[width=0.8\columnwidth]{images/metric_conceptual.pdf} }}%
    \qquad
    \subfloat[\label{fig:metric_concept_novel}]{{\includegraphics[width=0.8\columnwidth]{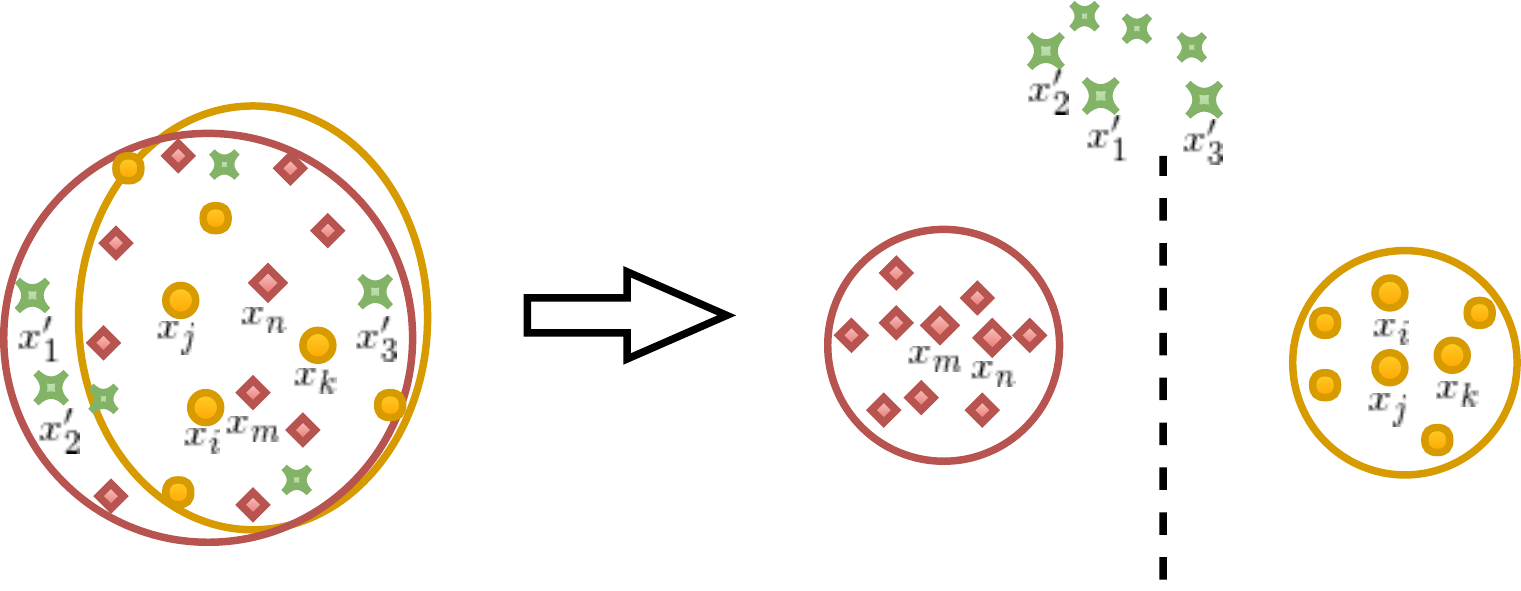} }}%
    \caption{(a) Two similar digits 1 and 7. (b) Illustration of local class cohesion assumption and novel class detection under it.}%
    \label{fig:intro_examples}%
\end{figure}

In this paper, we focus on another key problem called \emph{concept evolution}. Here, instances from previously unobserved classes (called novel classes) may occur along the stream. For example, images associated with classes for which the current classifier is not trained may appear along the stream during evaluation. If the classifier fails to account for the emerging classes, its performance would degrade. \looseness=-1
% In real-world data streams, such as image classification, text classification or intrusion detection, the number of classes is not fixed, and concept evolution may occur at any time. If the stream  classifier fails to address concept evolution timely, it might misclassify instances from a new class as existing classes, leading to degradation of classification performance. 

Recent studies \cite{masud2011,haque2016} have leveraged unsupervised clustering mechanisms, such as K-Means, over the observed feature space for detecting instances from novel classes. Here, clusters of instances represent regions in feature space containing instances of the same class label. Any instance that occurs outside the decision boundary of these clusters is referred as an \emph{outlier}.
Instances from a novel class are detected based on the density of outliers in feature space.
Such detection mechanisms rely on the existence of strong \emph{cohesion} among instances from the same class and large \emph{separation} among instances from different classes in observed feature space~\cite{masud2011}. We refer to this as global class cohesion and separation assumption respectively.
% Those outliers forming an isolated dense cluster are detected as coming from a novel class. 
% However, all these existing approaches depend on a strong universal closeness assumption~\cite{masud2011}, i.e., all instances associated with the same class in feature space are closer to each other (i.e., cohesion) compared to instances associated with different classes (i.e., separation). 
However, such a property may not be true in many real-world scenarios. A typical example is a handwritten digit recognition application where images of digit ``1" may look very similar to those of digit ``7", as shown in Figure~\ref{fig:digit_similarity}.
% Overall, images of these two classes may appear close to each other in feature space, along with images of other classes as well.
% Similarly, a news article about the takeover of a company can be associated with a class named ``business''. However, a large portion of its content may contain words related to ``technology'' class.
% is a ``business" news, but a large portion of its content may be an introduction of ``technology" patents owned by the company.
% Another example is the topic classification of online news feed stream where many common words could be used in news stories of different topics. For example, a news story about the takeover of a technical company is a "business" news, but a large portion of its content may be an introduction of patents owned by this company. 
In such cases, existing approaches fail to detect novel classes (e.g., class "7") from existing classes (e.g., class "1").
% and degrade in performance.
Alternately, Mu et al.\cite{mu2017streaming} proposed a framework which dynamically maintains two kinds of low-dimensional matrix sketches that approximate the original information along the stream. Novel class detection is performed using the encoded information in a low-dimensional space. Yet, this approach may be ineffective since the detection and dimensionality transformation are unrelated processes.
% to detect novel classes. 

In this paper, we address previous challenges by proposing a framework that can perform label prediction under concept evolution, called \sysname{} (\underline{C}onvolutional open-world multi-task image \underline{S}tream classifier with \underline{I}ntrinsic similarity \underline{M}etrics). 
The main goal is to transform the observed raw images into a latent feature space such that the classifier loss is minimized from achieving cohesion among instances belonging to the same class and separation of instances belonging to different classes. We achieve this by learning a latent feature space suitable for novel class detection. Particularly, we jointly train three main form of data transformation. First is a set of convolution layers to learn high-level features of images. These features are then transformed into another latent feature space using metric learning mechanisms~\cite{metric_survey} so that cohesion and separation properties can be distinctly achieved. We then employ novel class detection mechanism within this transformed feature space for data classification. 
% first using a convolution layer to transform images  jointly leveraging metric learning mechanisms~\cite{metric_survey} with a novel class detection mechanism to form a multi-task learning scenario. 
% This relaxes the assumption of class cohesion in observed feature space.
% This relaxes the assumption of global class cohesion and separation made by previous approaches.
% into the local class cohesion assumption.
% The main idea is to fuse metric learning together with novel class detection into stream classification via multi-task learning. Our framework is semi-supervised since we only request true labels from a small number of instances along the stream. Unlike previous approaches, \sysname{} relaxes the universal closeness assumption to the instance closeness assumption, i.e., any instance associated with a class in feature space is close to one or more different instances from the same class. Note that we make no assumption about the relationship among classes. 
For example, suppose we have three sets of instances, as shown in Figure~\ref{fig:metric_concept_novel}. Here. $\lbrace\bm{x}_i, \bm{x}_j, \bm{x}_k\rbrace$, $\lbrace\bm{x}_m, \bm{x}_n \rbrace$ and $\lbrace \bm{x}_1',\bm{x}_2',\bm{x}_3' \rbrace$ are instances associated with class A, class B and class C respectively. Considering class A and B, in observed feature space.
% $\bm{x}_i$ should be close to $\bm{x}_j$ and $\bm{x}_k$, compared to $\bm{x}_m$ and $\bm{x}_n$. 
$\bm{x}_i$ should be close to either $\bm{x}_j$ or $\bm{x}_k$, while $\bm{x}_m$ should be close to $\bm{x}_n$. However, since no assumption is made on the cohesion in $\lbrace\bm{x}_i, \bm{x}_j, \bm{x}_k\rbrace$ and $\lbrace\bm{x}_m, \bm{x}_n \rbrace$, $||\bm{x}_i-\bm{x}_j||_2 \gg ||\bm{x}_i-\bm{x}_m||_2$ is possible. Using \sysname, we aim to transform the instances to an appropriate latent feature space so as to satisfy the closeness constraint for novel class detection, as illustrated in Figure~\ref{fig:metric_concept_novel}. Here, we first obtain a high-level feature embedding with the aid of convolution layers and then learn a latent feature space from instances of class A and B, while class C is the novel class. In the observed feature space, $\lbrace \bm{x}_1',\bm{x}_2',\bm{x}_3' \rbrace$ are close to instances of class A and are relatively far from each other. After the transformation, instances of each class form a dense cluster and are separated with a large margin, making novel class detection possible.
% Due to the more practical instance closeness assumption, \sysname{} outperforms many existing approaches on complex real-world tasks. We evaluate this framework on real-world datasets, and compare its performance with competing baselines and state-of-the-art novel class detection systems.
 
The contributions of this paper are as follows:
 \begin{itemize}
 \item We present a semi-supervised framework called \sysname{} that addresses the challenges of classification and concept evolution on high-dimensional real-world image streams.
 \item We propose a unified multi-task classifier that jointly performs metric learning, stream classification, and novel class detection. 
%  It is trained by fusing metric learning and novel class detection into stream classification via multi-task learning.
 \item We empirically evaluate \sysname{} on real-world datasets, and compare its results with existing state-of-the-art novel class detection systems. We also study the effectiveness of the proposed feature transformation by comparing its performance with other metric learning approaches.
\end{itemize}

The rest of this paper is organized as follows. In Section~\ref{Sec:background}, we present a brief background on metric learning and stream classification. We then formally define the problem and present its challenges in Section~\ref{Sec:preliminaries}, before detailing the proposed solution in Section~\ref{Sec:approach}.
In Section~\ref{Sec:evaluation}, we present the results of our empirical evaluation and finally conclude in Section~\ref{Sec:conclusion}.

\section{Related Works}
\label{Sec:background}
\subsection{Metric Learning}
\label{sec:background-metric}
% The primary contribution of this paper concerns the choice of data transformation for novel class detection.
Distance-based metric learning~\cite{metric_survey} plays a significant role in pattern recognition. Studies\cite{hu2014, weinberger2009} have successfully applied this to address complex classification tasks in the real world. 
Following the early work of Xing et al.\cite{xing2003}, the goal of metric learning is to learn a distance-metric that minimizes the distance between similar examples and maximizes that between dissimilar examples. A distance-metric is usually represented as either an \textit{Explicit Metric Function} (EMF) or an \textit{Implicit Metric Function} (IMF). 

\subsubsection{Explicit Metric Function}
The explicit metric function can be viewed as a linear/non-linear embedding function that maps examples in the original feature space into a new transformed feature space \cite{davis2007, goldberger2005, guillaumin2009, weinberger2009}. A common closed-form linear EMF is the Mahalanobis-like distance $D_{M}^{2}(\bm{x},\bm{y}) = (\bm{x} - \bm{y})^{T}M(\bm{x} - \bm{y})$ \cite{xing2003}, where $M$ is a positive semi-definite (PSD) matrix satisfying the training constraints. This Mahalanobis-like distance introduces a linear transformation which maps $\bm{x}$ to $\bm{x'}=L\bm{x}$ with $M=L^TL$. However, the simplicity of linear EMF limits its application on complex tasks.
%     and therefore the non-linear EMF, which is usually learned by generalizing the Euclidean distance with a non-linear transformation $\phi$, is proposed.
To address this issue, non-linear EMF, which is usually learned by generalizing the Euclidean distance with a non-linear transformation $\phi$, is proposed.
In this case, the distance measure becomes $d_{\phi}(\bm{x}, \bm{y})=||\phi(\bm{x})-\phi(\bm{y})||_2$.
    
\subsubsection{Implicit Metric Function}
In contrast to explicit metric functions, it is inconvenient to obtain an explicit expression of the transformed embedding space for implicit metric functions. Many techniques have been adopted to learn an IMF and the widely accepted method is the kernel approach. For an input feature space $\mathcal{H}$, a kernel $\mathcal{K}: \mathcal{H}\times\mathcal{H}\rightarrow \mathcal{R}^+$ is a positive-definite function that are bivariate measures of similarity based on the inner product between samples embedded in a Hilbert space. Although implicit metric functions work well in some applications like clustering, 
%   constructing a kernel matrix is costly and it is hard to apply the learned metric function to prediction on future unseen instances. 
constructing a kernel matrix is computationally expensive and applying the learned IMF for future predictions is difficult.
In this paper, we focus on the non-linear EMF and present a novel approach that learns a high-quality metric via multi-task learning.

% \subsection{Multi-Task Learning}
% \textit{Multi-Task Learning} (MTL) is a learning paradigm which aims to leverage useful information contained in multiple related tasks to help improve the generalization performance of all tasks. Formally, given $m$ learning tasks $\lbrace \mathcal{T}_i\rbrace_{i=1}^m$ where all the tasks or a subset of them are related, \textit{multi-task} learning aims to help improve the learning of a model for $\mathcal{T}_i$ by using the knowledge contained in the m tasks~\cite{ZhangY17aa}. A particular type of multi-task learning assumes that different tasks share some common parameters but have their own adapted parameters to generate outputs. For example, in neural network, the common parameters are represented by shared hidden layers while those adapted parameters are represented by additional individual layers specific to each task. 

\subsection{Stream Classification}
\label{subsec:sc}
% Stream classification with novel class detection has been widely studied in recent years. 
A \textit{novel class} at time $t>0$ is defined as a class label whose associated instances have never been observed along the stream until time $t$. Therefore, a classifier is never trained or updated using instances associated with this class. Studies typically aim to detect such novel class instances and reactively adapt the classifier for better performance.
Previous studies in this direction \cite{masud2011,haque2016} have developed frameworks that leverage an unsupervised mechanism called Q-NSC for novel class detection. It uses the clusters resulting from K-Means to detect outliers, which are then analyzed based on density to detect novel classes. Alternatively, the framework by Mu et al.\cite{mu2017streaming} uses low dimensional matrix sketches~\cite{mu2017streaming} by leveraging frequent directions~\cite{GhashamiLPW16} to detect novel class. 
Furthermore, all these approaches use a user-defined threshold to identify instances from novel classes along the stream. Unlike them, we employ a multi-task learning technique with online threshold evaluation for novel class detection.

\section{Preliminaries}
\label{Sec:preliminaries}
% The vast majority of existing data stream classification techniques with novel class detection (NCD) module discussed in Section~\ref{Sec:background} make two strong assumptions that are often violated in the real-world. First, they depend on the strong cohesion property of same-class instances for correct predictions. Second, they also depend on the strong separation property in observed feature space among instances from different classes for novel class detection. 
% assumed that examples of any novel class are far from the known classes in feature space. 
% This strong assumption is often violated in real-world scenarios since instances from the novel class might be similar to that from the known classes. 
% We address this issue by first formally defining the problem as follows.

In this section, we formally define the problem and list the associated challenges we address in this paper.

\subsection{Problem Statement}
Given a training dataset $D=\lbrace(\bm{x_i}, y_i)\rbrace_{i=1}^m$, where $\bm{x_i}\in \mathcal{R}^d$ is a training instance and $y_i \in \mathcal{Y}=\lbrace1,2,\ldots, c\rbrace$ is the associated class label, and a non-stationary streaming data $S=\lbrace(\bm{x_t}, y_t)\rbrace_{t=1}^\infty$, where $\bm{x_t}\in\mathcal{R}^d$ and $y_t\in\mathcal{Y}'=\lbrace 1,2,\ldots,c, c+1, \ldots, c'\rbrace$ ($c'>c$), the goal is to learn a model $f$ (initially with $D$) such that $f(\bm{x_t}) \rightarrow \mathcal{Y}'$. For every incoming instance in the data stream, $f$ will determine whether it belongs to an unknown (also referred as novel) or an existing class. 
% If yes, this instance is placed in a buffer $\mathcal{B}$ which stores candidates of previously unseen classes; Otherwise, $f$ produces a class prediction. Once the number of candidates in $\mathcal{B}$ reaches the buffer size limit, they will be used to update model $f$.
Note that for any two arbitrary classes $c_m, c_n \in \mathcal{Y}'$ ($c_m\ne c_n$), if $\lbrace\bm{x_i}, \bm{x_j}\rbrace \in c_m $ and $\lbrace\bm{x_k}\rbrace \in c_n$, it is possible that $||\bm{x_i}-\bm{x_k}||_2<||\bm{x_i}-\bm{x_j}||_2$. The overall aim of the task is to maintain high classification accuracy along a data stream where instances from novel classes may occur over time. 
% and detect instances from novel classes as precisely as possible.

\subsection{Challenge}
We assume that a data stream is non-stationary and consider a practical scenario where instances belonging to the same class may be further away from each other than instances from other classes in observed feature space. This introduces three main challenges:
\begin{itemize}
\item Model $f$ has to capture instance similarity and dissimilarity correctly in a latent feature space suitable for class discrimination. It means that model $f$ should internally learn a similarity metric capable of classifying high-dimensional data patterns with little external information.
\item Due to the unbounded length of a data stream, model $f$ could only be trained with a limited amount of training data at a given time and yet should predict well over long periods of time.
% but should predict well on a vast amount of future test instances.
\item Since novel classes could appear continuously in a data stream, model $f$ needs to detect the emergence of novel classes while requiring a small amount of truth-value for the model update when necessary.
\end{itemize}

\section{The Proposed Approach}
\label{Sec:approach}
\begin{table*}[t] %\vspace*{-0.2cm}

\centering
\resizebox{0.8\textwidth}{!}{%
\begin{tabular}{ |l l|}
\hline
$\mathcal{S}$: Stream data & $\mathcal{B}$: Novel class candidate buffer \\
$S_{\mathcal{B}}$: The maximum size of $\mathcal{B}$ & $\mathcal{D}$: Data storage\\
$S_{\mathcal{D}}$: The maximum size of each class in $\mathcal{D}$ & $\mathcal{T}_{novel}$: Confidence threshold for novel class detection \\
$\mathcal{T}_{\mathcal{D}}$: Confidence threshold for updating $\mathcal{D}$ & $D$: Initial training data in warm-up phase\\
$\mathcal{Y}$: Label set of initial training data & $\mathcal{Y}'$: Open set of possible labels in $\mathcal{S}$\\
$\bm{x}$: d-dimensional features & $y\in \mathcal{Y}'$: Class label of a data instance \\
$\mathit{f}$: Open-world classifier & $\mathcal{P}(\bm{x})$: Prediction confidence for $\bm{x}$ using $\mathit{f}$ \\
$\tilde{y}$: Final predicted label of a data instance & $\gamma$: Significance level of margin for triplet loss \\
$W_i$: Weights associated with class $c_i$ in 1-vs-rest layer & $\mathcal{Y}_{\mathcal{D}}$: Label set of data storage $\mathcal{D}$\\
$\hat{y}$: Estimated label of a data instance & $S_{update}$: Minimum number of instances of a class in $\mathcal{D}$ for classifier update\\
$S_{\emph{mini}}$: Mini-batch size for MBGD &$n_{e}$: number of epochs \\
\hline
\end{tabular}
}
\caption{\textbf{Frequently used symbols}}
\label{tab:symbols}
\end{table*}

% In this section, we describe the overview of the proposed framework \sysname{}, present its constitution and discuss the details of each component. 
In this section, we describe our proposed framework \sysname{} for stream classification by first presenting an overview of \sysname{} and then discussing each component in details.

\subsection{Overview}
\label{subsection: overview}
% In this paper, we propose a simple but efficient solution for single stream classification which addresses the above challenges  by fusing metric learning and novel class detection into stream classification based on multi-task learning. We refer to this approach as \sysname{} (efficient single Stream classification with Built-In Metrics via multi-task learning).
\begin{figure}[t]
\centering
\includegraphics[width=\columnwidth]{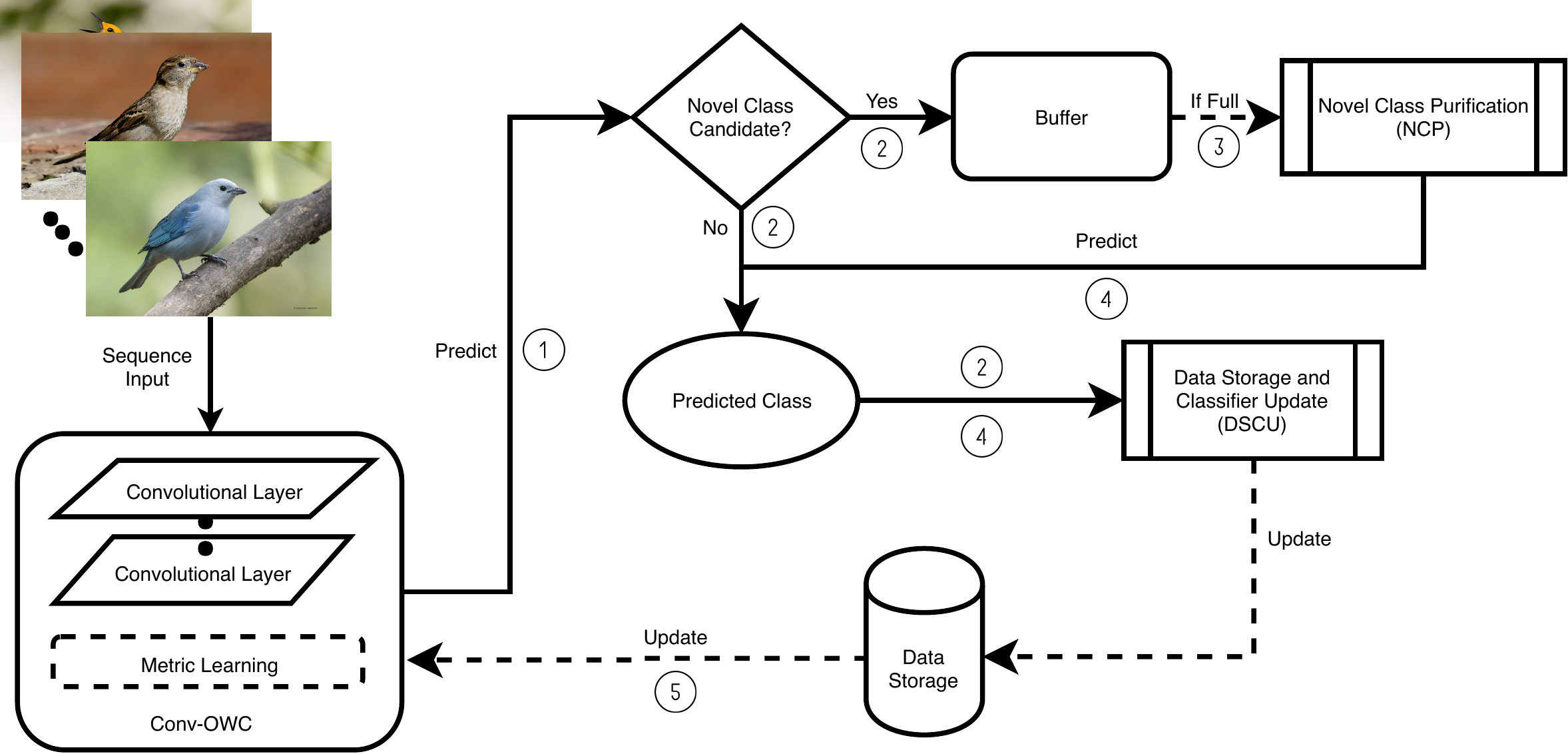}
\caption{Overview of framework. (Numbers in circles represent the execution order of modules in \sysname{})}
\label{fig:framework}
\end{figure}

To determine instances belonging to a novel class over a data stream, a typical technique requires sufficient amount of instances to have a density beyond a user-given threshold over the observed feature space.
% The existence of a novel class is determined using a threshold function over instance density in feature space. 
As a reaction, the classifier is trained to predict over classes that include the detected novel class. We use this mechanism by first transforming the observed feature space into appropriate latent space utilizing a combination of convolution and a unique distance-based metric to identify potential novel class instances and then updating the classifier correspondingly. To achieve this goal, we propose a framework called \sysname{}.

Figure~\ref{fig:framework} illustrates the core components and classification process in \sysname{}. It has five main modules, i.e., \textit{Convolutional Open-World Classifier (Conv-OWC)}, \textit{Metric Learning (ML)}, \textit{Classification}, \textit{Novel Class Purification (NCP)}, and \textit{Data Storage and Classifier Update (DSCU)}. 
% Particularly, an Open-World Classifier~\cite{bendale2015} is trained only on partially observed classes over a dataset.
% and should be able to deal with myriad of unseen class labels during evaluation. 
At first, the classifier is trained on an initial set of instances in $D$. 
For any new instance $\bm{x}$ arriving in $\mathcal{S}$, its estimated label $\hat{y}$ is the maximum likelihood prediction from the Conv-OWC. The convolutional layers in Conv-OWC identifies the edges of the incoming instance $\bm{x}$ and retrieves a conceptual representation of $\bm{x}$ which is then sent to metric embedding layer to produce a high-level embedding used for classification. 
If the prediction result indicates that $\bm{x}$ is not a potential candidate from any novel class, i.e., $\hat{y}\ne -1$, then the final predicted label $\tilde{y}$ for $\bm{x}$ is $\hat{y}$ , i.e., $\tilde{y}=\hat{y}$; Otherwise, $\bm{x}$ is temporarily stored in the candidate buffer $\mathcal{B}$.

As new instances arrive in $\mathcal{S}$, the Novel Class Purification (NCP) module monitors the size of $\mathcal{B}$. Once the buffer $\mathcal{B}$ is full, the NCP module detects the existence of any instance from unknown classes in $\mathcal{B}$. Moreover, it separates them from known class instances that may be present due to noise in the stream. These instances are then used to update the data storage $\mathcal{D}$ and a new model is trained on the updated $\mathcal{D}$ if the update condition is satisfied. Algorithm~\ref{alg:\sysname{}} illustrate the details of classification and novel class detection process in \sysname{}.
% The details of modules in \sysname{} are discussed as below. 

%%%%%%%%%%% Algorithm-approach %%%%%%%%%%%%%
\begin{algorithm}
\caption{\textbf{\sysname{}: Stream Classification}}
\label{alg:\sysname{}}
\begin{algorithmic}[1]

\REQUIRE $\mathcal{S}$ - Stream data; $S_{\mathcal{B}}$ - The maximum size of $\mathcal{B}$; $\mathcal{T}_{\mathcal{D}}$ - Confidence threshold for updating $\mathcal{D}$; $D$ - Initial training data in warm-up phase; 
\ENSURE Label $\tilde{y}$ predicted on $\mathcal{S}$ data.
\STATE Learn an initial model $\mathit{f}$ from $D$ by solving the optimization problem. (Eq.~\ref{eq:overall_optimize}) \\
\REPEAT
	\STATE Receive a new instance $\bm{x}$. \\
    \STATE Predict label $\hat{y}$ for $\bm{x}$ using $\mathit{f}$ according to Eq.~\ref{eq:prediction}
    \IF{$\hat{y}= -1$}
    \STATE Store $\bm{x}$ in the candidate buffer $\mathcal{B}$.
    \ELSE
    \STATE $\tilde{y}\leftarrow \hat{y}$
    \ENDIF
    \IF{size($\mathcal{B}$)$\ge S_{\mathcal{B}}$}
    \STATE Check for occurrence of any novel class in data using \textit{DetectNovel} (Algorithm~\ref{alg:ECD})
    \IF{\textit{DetectNovel} returns \textit{True}}
    \IF{Update-Condition(Section~\ref{sec:update}) Satisfied}
    \STATE Retrain $\mathit{f}$ with $\mathcal{D}'$ (a subset of $\mathcal{D}$) (Section~\ref{sec:update}).
    \ENDIF
    \ENDIF
    \ENDIF
    \IF{$\mathcal{P}(\bm{x})> \mathcal{T}_{\mathcal{D}}$}
    \STATE Update $\mathcal{D}$ using $(\bm{x}, \tilde{y})$ (Section~\ref{sec:update}).
    \ENDIF
\UNTIL $\mathcal{S}$ exits

\end{algorithmic}
\end{algorithm}
%%%%%%%%%%%%%%%%%%%%%%%%%%%%%%%%%%%%%%%%%%%
\subsection{Metric Learning (ML)}
\label{sec:ML}
% As discussed in Section~\ref{sec:background-metric}, 
A high-quality similarity metric is critical to the performance of both classification and novel class detection. Let $\lbrace(\bm{x}_1^t, y_1^t),\ldots,(\bm{x}_n^t, y_n^t)\rbrace \in \mathcal{R}^{d\times C_t}$ be all training data in $\mathcal{D}$ at time $t$, where $C_t=\lbrace1,\ldots,k\rbrace$ denotes $k$ different classes. Our goal is to find an explicit metric function (EMF) represented by $\phi(\bm{x})$ that transforms an instance $\bm{x}$ into a feature space $\mathcal{R}^{d'}$. Here, the transformation is such that the squared Euclidean distance between any pair of instances of the same class, independent of their locations in original space $\mathcal{R}^d$, is small and the squared Euclidean distance between any pair of instances from different classes is large. However, instead of considering only a pair of instances at a time, we focus on triplets. \looseness=-1

\begin{definition}[Triplet]
\label{def:triplet}
A triplet $(\bm{x}^a, \bm{x}^p, \bm{x}^n)$ is a group of three instances where $\bm{x}^a$ (anchor instance) is similar to $\bm{x}^p$ (positive instance) but is dissimilar to $\bm{x}^n$ (negative instance).
\end{definition}
% We believe triplet-based loss is more suitable for our problem. The motivation is that it not only encourages all instances of one class to be projected onto a single point in the embedding space but also enforces a margin between each pair of instances from one class to all other classes. Therefore, the instances from one class would live on a manifold and are still discriminable to other classes. 
Introduced in~\cite{SchroffKP15}, the triplet-based loss is more suitable for our problem since it encourages all instances of one class to be projected onto a single point in the embedding space while enforcing a small constant margin between each pair of instances from one class to all other classes. 
Unfortunately, this kind of loss attempts to over-compress instances from same class to produce a constant margin leading to overfitting. This margin makes novel class detection difficult. Moreover, it also requires a large volume of training data which is not available in stream applications. Observing these shortcomings, we appeal to a novel \textit{Triplet Loss} function based on triplets defined in Definition~\ref{def:triplet}.

\subsubsection{Triplet Loss}

Let $\mathbb{D}$ be a given training set that contains $M$ triplets and $(\bm{x}_i^a, \bm{x}_i^p, \bm{x_i^n)}$ be the $i^{th}$ triplet in $\mathbb{D}$. The EMF $\phi(\bm{x})$ embeds an instance $\bm{x}$ into a $d'$-dimensional Euclidean space. After embedding, we expect the Euclidean distance between $\bm{x}_i^a$ and $\bm{x}_i^n$ to be at least $\mathrm{e}^{\gamma}$ ($\gamma \ge 1$) times the distance between $\bm{x}_i^a$ and $\bm{x}_i^p$. Formally,
% \begin{equation}
% \frac{||\phi(\bm{x}_i^a)-\phi(\bm{x}_i^n)||_2}{||\phi(\bm{x}_i^a)-\phi(\bm{x}_i^p)||_2} \ge \mathrm{e}^{\gamma}
% \end{equation}
% After smoothing, it becomes
\begin{equation}
\label{eq:ratio}
\frac{||\phi(\bm{x}_i^a)-\phi(\bm{x}_i^n)||_2+1}{||\phi(\bm{x}_i^a)-\phi(\bm{x}_i^p)||_2+1} \ge \mathrm{e}^{\gamma}
\end{equation}
where $1$ is added as a smoothing factor. The resulting triplet loss $\mathcal{L}_{triplet}$ is
\begin{equation}
\begin{split}
\mathcal{L}_{triplet} =& \frac{1}{M}\sum\limits_{i=1}^M\bigg[\log(||\phi(\bm{x}_i^a)-\phi(\bm{x}_i^p)||_2+1)+\gamma \\ &-
\log(||\phi(\bm{x}_i^a)-\phi(\bm{x}_i^n)||_2+1)\bigg]_{+}\\
\end{split}
\label{eq: triplet_loss}
\end{equation}
Note that smoothing introduces an implicit constraint that at least one of $||\phi(\bm{x}_i^a)-\phi(\bm{x}_i^n)||_2$ and $||\phi(\bm{x}_i^a)-\phi(\bm{x}_i^p)||_2$ should be much greater than $1$; Otherwise, the ratio computed by Eq.~\ref{eq:ratio} would be close to 1 and the inequality is unsatisfied. 

The motivation of introducing $\mathcal{L}_{triplet}$ is that it pushes different classes further away from each other by introducing a larger instance-sensitive margin, since $||\phi(\bm{x}_i^a)-\phi(\bm{x}_i^n)||_2-||\phi(\bm{x}_i^a)-\phi(\bm{x}_i^p)||_2 \approx (\mathrm{e}^\gamma-1) ||\phi(\bm{x}_i^a)-\phi(\bm{x}_i^p)||_2$.

We want to minimize the triplet loss $\mathcal{L}_{triplet}$ but constrain the learned embedding on a $d'$-dimensional unit sphere. So the optimization problem for metric learning is 
\begin{equation}
\label{eq:optimization_problem}
\begin{aligned}
& \underset{\phi}{\text{minimize}}
& & \mathcal{L}_{triplet} \\
& \text{subject to}
& & ||\phi(\bm{x_i^*})||_2=1, \forall \bm{x_i^*} \in \mathbb{D}.\\
\end{aligned}
\end{equation}
Here $\bm{x_i^*}$ denotes any anchor, positive or negative instances in $\mathbb{D}$, according to the definition. Any non-linear function can be utilized as $\phi$ in the optimization problem. So, we choose to represent $\phi$ as a convolutional neural network with a single fully-connected hidden layer of $\mathtt{n}$ units for simplicity.
\begin{figure}[t]
\centering
\includegraphics[width=\columnwidth]{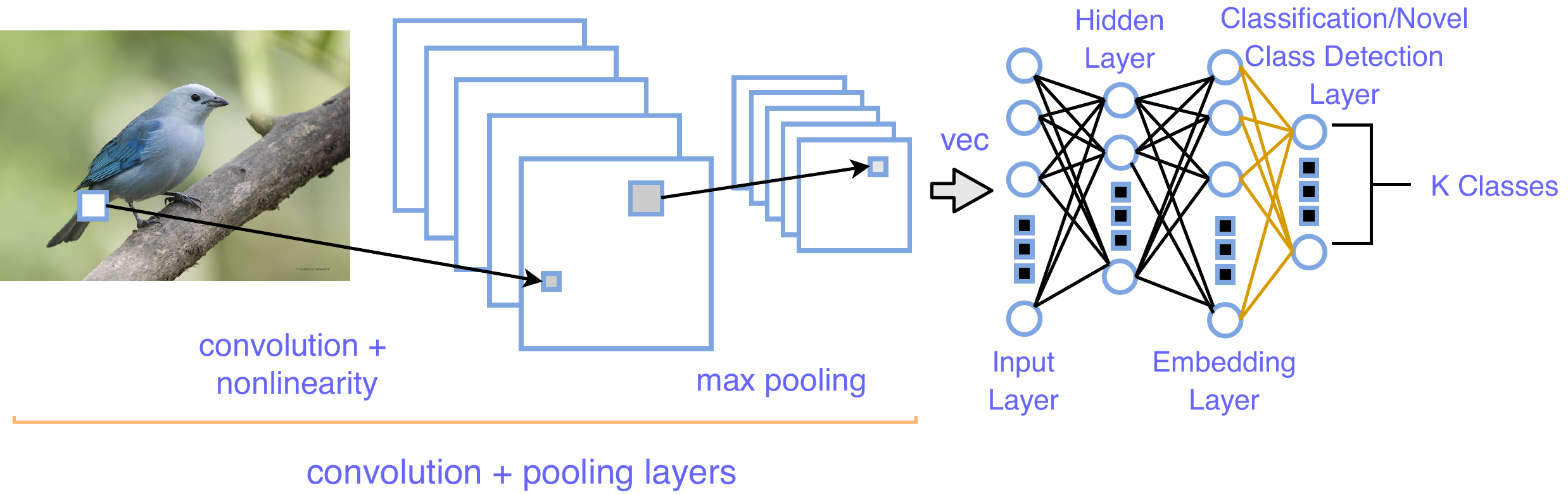}
\caption{Structure of Convolutional Open-World Classifier.}
\label{fig:open-word-classifier}
\end{figure}

The choice of triplets used for metric training is critical to the quality of learned metric. However, generating all possible combinations would result in a large number of triplets that are easily satisfied (i.e. fulfill the constraint in Eq.~\ref{eq: triplet_loss}) that do not contribute to the training process. This may result in slow convergence. Therefore, it is crucial to select \emph{hard} triplets that continuously contribute to improving the model. Here, we use the term ``hard'' to indicate a positive loss.

\subsubsection{Triplets Selection}
A triplet can be generated by first selecting an ``anchor" class $c_a$ and a ``negative" class $c_n$ ($c_a\ne c_n$) from $\mathcal{D}$ and then choose two different instances $\bm{x_i^a}$ and $\bm{x_i^p}$ from $c_a$, and one instance $\bm{x_i^n}$ from $c_n$. As mentioned above, we want to generate "hard" triplets for fast convergence. This means that, given $\bm{x_i^a}$, we want to select an $\bm{x_i^p}$ (\textit{hard positive}) such that $\argmin_{\bm{x_i^p} }||\phi(\bm{x_i^a})-\phi(\bm{x_i^p})||_2^2$ and an $\bm{x_i^n}$ (\textit{hard negative}) such that $\argmax_{\bm{x_i^n} }||\phi(\bm{x_i^a})-\phi(\bm{x_i^n})||_2^2$. However, computing the $\argmin$ and $\argmax$ across $\mathcal{D}$ is computationally intractable due to a large search space. Therefore, we aim to generate triplets in an online fashion. We focus on a mini-batch approach consisting of a subset of instances randomly sampled from $\mathcal{D}$ at each step. By applying mini-batch gradient descent (MBGD) approach for minimizing $\mathcal{L}_{triplet}$, we transform the instances to the embedding space. Then we compute the $\argmax$ and $\argmin$ within that mini-batch to generate desired triplets. The motivation behind this decision is to provide hard triplets to the model at any stage during its training to continuously improve the learned embedding. 

\subsection{Convolutional Open-World Classifier (Conv-OWC)}
Due to the important role of a high-quality metric in both classification and novel class detection, we choose to fuse metric learning and novel class detection into classification. Hence, we propose a novel classifier referred as \textit{Convolutional Open-World Classifier (Conv-OWC)} that performs all these tasks jointly. Figure~\ref{fig:open-word-classifier} illustrates the structure of Conv-OWC in \sysname{}. The \textit{Convolutional layer, Max-Pooling layer, Input Layer}, \textit{Hidden Layer} and \textit{Embedding Layer} learns the metric $\phi$. 
% Following~\cite{DBLP:conf/emnlp/ShuXL17}, 
In contrast to traditional multi-class classifiers that typically use softmax as the final output layer, we build a 1-vs-rest layer (\textit{Classification/Novel Detection Layer}) containing $K$ sigmoid functions for $K$ classes, following~\cite{DBLP:conf/emnlp/ShuXL17}. For $i^{th}$ sigmoid function corresponding to class $c_i$, Conv-OWC takes all examples with label $y=c_i$ as positive examples and the rest with $y\ne c_i$ as negative examples. Let $\mathcal{L}_{class}$ denotes the loss introduced by the 1-vs-rest layer. It is the average Binary Classification Error (BCE) of $K$ sigmoid functions on the training data $\mathcal{D}$. Formally, the loss is given by:

\begin{equation}
\begin{split}
\mathcal{L}_{class} =& \frac{1}{Kn}\sum\limits_{i=1}^K\sum\limits_{j=1}^n [-\mathbb{I}(y_j = c_i)\log\mathcal{P}(y_j=c_i)\\
&-\mathbb{I}(y_j\ne c_i)\log(1-\mathcal{P}(y_j=c_i))]\\
\end{split}
\end{equation}

Unlike~\cite{DBLP:conf/emnlp/ShuXL17}, we do not optimize $\mathcal{L}_{class}$ on its own. Instead, we optimize it with the triplet loss $\mathcal{L}_{triplet}$. Thus the tasks of metric learning, classification and novel class detection are learned jointly in Conv-OWC. The resulting objective function for multi-task optimization, denoted as $\mathcal{L}_{overall}$, is 
% \begin{equation}
% \begin{split}
% \mathcal{L}_{overall} =& \sum\limits_{i=1}^K\sum\limits_{j=1}^n [-\mathbb{I}(y_j = c_i)\log\mathcal{P}(y_j=c_i)\\
% &-\mathbb{I}(y_j\ne c_i)\log(1-\mathcal{P}(y_j=c_i))]\\
% &+\beta \cdot \{ \frac{1}{M}\sum\limits_{i=1}^M[\log(||\phi(\bm{x}_i^a)-\phi(\bm{x}_i^p)||_2+1) \\
% & + \gamma - \log(||\phi(\bm{x}_i^a)-\phi(\bm{x}_i^n)||_2+1)]_{+}\} \\
% \end{split}
% \end{equation}
\begin{equation}
\begin{split}
\mathcal{L}_{overall} =& \sum\limits_{j=1}^M \Bigg\{ \bigg(\frac{1}{3KM}\sum\limits_{i=1}^K \sum\limits_{*\in \{a, p ,n\}}\big[-\mathbb{I}(y_{\bm{x_j^*}} = c_i)\log\mathcal{P}(y_{\bm{x_j^*}}=c_i)\\
&-\mathbb{I}(y_{\bm{x_j^*}}\ne c_i)\log(1-\mathcal{P}(y_{\bm{x_j^*}}=c_i))\big]\bigg)\\
&+ \frac{\beta}{M}\bigg[\log(||\phi(\bm{x}_j^a)-\phi(\bm{x}_j^p)||_2+1) \\
& + \gamma - \log(||\phi(\bm{x}_j^a)-\phi(\bm{x}_j^n)||_2+1)\bigg]_{+} \Bigg\}\\
\end{split}
\end{equation}
where $\mathcal{P}(y_{\bm{x_j^*}}=c_i)=\sigma(W_i\phi(\bm{x_j^*})+b)$ ($W_i$ is the weight of $i^{th}$ class in 1-vs-rest layer), $\beta$ is a hyper-parameter that controls the importance of $\mathcal{L}_{triplet}$ in $\mathcal{L}_{overall}$ and $M$ is the number of triplets used for training.
The overall optimization problem is given by 
\begin{equation}
\begin{aligned}
& \underset{\phi, W_1, W_2, \ldots, W_K}{\text{minimize}}
& & \mathcal{L}_{overall} \\
& \text{subject to}
& & ||\phi(\bm{x_i^*})||_2=1, \forall \bm{x_i^*} \in \mathbb{D}.\\
\end{aligned}
\label{eq:overall_optimize}
\end{equation}

% aims to leverage useful information contained in multiple related tasks to help improve the generalization performance of all tasks. Formally, given $m$ learning tasks $\lbrace \mathcal{T}_i\rbrace_{i=1}^m$ where all the tasks or a subset of them are related, \textit{multi-task} learning aims to help improve the learning of a model for $\mathcal{T}_i$ by using the knowledge contained in the m tasks~\cite{ZhangY17aa}. A particular type of multi-task learning assumes that different tasks share some common parameters but have their own adapted parameters to generate outputs. For example, in neural network, the common parameters are represented by shared hidden layers while those adapted parameters are represented by additional individual layers specific to each task. 
By optimizing $\mathcal{L}_{overall}$, the knowledge learned via metric learning helps improve the generalization performance of classification and vice versa. This information transfer in $\mathcal{L}_{overall}$ is critical in stream applications where a limited amount of labeled training data is available.

\subsection{Classification}
Suppose $\mathit{f}$ denotes the convolutional open-world classifier and $\breve{y}$ is the prediction label generated by $f$, for every incoming instance $\bm{x}$, the prediction probability $\mathcal{P}(\breve{y}=c_i|\bm{x})$ of class $c_i$ is computed by $\mathcal{P}(\breve{y}=c_i|\bm{x})=\sigma(W_i\phi(\bm{x})+b)$. However, before making a decision on the predicted label of instance $\bm{x}$, we need to determine the threshold $\mathcal{T}_{novel}$ for novel class detection. 

Due to the non-stationary nature of stream, it is inappropriate to manually set a threshold and expect it work well along the stream. This indicates that the threshold for novel class detection should be determined automatically based on current stream property. To obtain a better $\mathcal{T}_{novel}$, we use the idea of one-sided confidence bound in statistics.

Assume the predicted probabilities $\mathcal{P}(\breve{y}=c_i|\bm{x})$ for all data of each class $c_i$ in a training dataset $\mathcal{D}$ follow a Gaussian distribution with unknown mean and unknown variance. A good statistic for confidence threshold is the average prediction probability of training data, i.e., $\bar{\mathcal{P}}(c_i) = \frac{1}{||\mathcal{D}_{i}||}\sum\limits_{\mathcal{D}_{i}}\mathcal{P}(\breve{y} = c_i|\bm{x}\in \mathcal{D}_i)$, where $\mathcal{D}_i=\lbrace(\bm{x_j}, y_j=c_i), \forall \bm{x}_j \in \mathcal{D}\rbrace$. $\bar{\mathcal{P}}(c_i)$ has a $t$ distribution with $||D_i||-1$ degrees of freedom. The desired $\mathcal{T}_{novel}$ for class $c_i$ is the $100(1-\alpha)\%$ confidence lower bound of $\bar{\mathcal{P}}(c_i)$ given by
\begin{equation}
\mathcal{T}_{novel}(c_i)= \bar{\mathcal{P}}(c_i) - t_{\alpha, ||\mathcal{D}_i||-1}S_{c_i}/\sqrt[]{||\mathcal{D}_i||}
\end{equation}
Here, $S_{c_i}$ is the sample standard deviation of $\lbrace\mathcal{P}(\breve{y}=c_i|\bm{x}), \forall \bm{x} \in \mathcal{D}_i\rbrace$.

Once we have the threshold $\mathcal{T}_{novel}$, classification is trivial. For the $i^{th}$ sigmoid function, we check if the predicted probability $\mathcal{P}(\breve{y}=c_i|\bm{x})$ is less than the NCD threshold $\mathcal{T}_{novel}(c_i)$. If the predicted probabilities of all classes are less than their corresponding thresholds for $\bm{x}$, then $\bm{x}$ is a candidate from a novel class. As a result, this instance is rejected (predicted as $-1$), and is temporarily stored in $\mathcal{B}$. Otherwise, its predicted class is the one with the highest probability. Formally, we have the following. \looseness=-1
\begin{equation}
\label{eq:prediction}
\hat{y} = 
\begin{cases}
-1 & \text{if }\mathcal{P}(\breve{y}=c_i|\bm{x}) < \mathcal{T}_{novel}(c_i), \\
& \forall c_i\in \mathcal{Y}_{\mathcal{D}}\\
\argmax\limits_{c_i\in\mathcal{Y}_{\mathcal{D}}} \mathcal{P}(\breve{y}=c_i|\bm{x}) & \text{otherwise}
\end{cases}
\end{equation}

Here $\hat{y}$ is the estimated label for an instance $\bm{x}$ and $\mathcal{Y}_{\mathcal{D}}$ is the label set of $\mathcal{D}$. If $\hat{y}\ne -1$, the final predicted label $\tilde{y}$ is the same as $\hat{y}$, i.e., $\tilde{y}=\hat{y}$; Otherwise, the prediction of $\tilde{y}$ for those instances with $\hat{y}=-1$ is left to the novel class purification module. \looseness=-1

\subsection{Novel Class Purification (NCP)}
Unlike many prior stream classifiers~\cite{masud2011, haque2016}, we make a more practical assumption that instances from a novel class might be similar to those from known classes in the observed feature space. Moreover, noise in streams may lead to false alarms. Hence, some instances from known classes might be incorrectly reported as coming from a novel class. Once the candidate-buffer $\mathcal{B}$ is full, the novel class purification module is invoked to filter novel class instances out from candidates in $\mathcal{B}$. It is done by following the steps below:
\begin{itemize}
\item Candidates in $\mathcal{B}$ is first transformed to the metric embedding space represented by $\phi$ and then \textit{DBSCAN}~\cite{ester1996} is performed on the transformed instances to achieve a set of clusters $\lbrace \mathcal{C}_1, \ldots, \mathcal{C}_m\rbrace$.
\item For each $\mathcal{C}_i$, we randomly sample out one instance from the cluster to request its true label and this true label would be the prediction label for all instances within this cluster. \looseness=-1

\end{itemize}
Here, DBSCAN is selected since it is unsupervised and does not have a strong constraint regarding cluster shape like K-Means. After being transformed into metric embedding space, instances from the same class tend to form a dense cluster. Although clusters of novel classes are separated from those of existing classes with a larger margin in \sysname{}, they are not sufficiently far away so that global separation assumption could hold. It is due to the lack of novel class information during the training of $\phi$. Those detection techniques based on the global separation assumption would simply fail in this case. However, the cohesion property of clusters indicates that instances within a cluster are semantically similar to each other. This builds the foundation of our proposed NCP. A formal description of the NCP is shown in Algorithm~\ref{alg:ECD}.

%%%%%%%%%%% Algorithm-approach %%%%%%%%%%%%%
\begin{algorithm}[t]
\caption{\textbf{DetectNovel}}
\label{alg:ECD}
\begin{algorithmic}[1]

\REQUIRE Candidate Buffer $\mathcal{B}$ 
\ENSURE True/False
\STATE $\mathbb{C}=\lbrace\mathcal{C}_1,\ldots,\mathcal{C}_m\rbrace$ $\leftarrow$ \textit{DBSCAN}($\mathcal{B}$)
\STATE $Novel\leftarrow False$
\FOR{$\mathcal{C}_i\in \mathbb{C}$}
\STATE Randomly sample a instance $\bm{x}_{c_i}\in \mathcal{C}_i$.
\STATE Request truth label $y_{c_i}$ of $\bm{x}_{c_i}$.
\IF{$y_{c_i}$ is unknown before}
\STATE	$Novel\leftarrow True$
\ENDIF
\FOR{$\bm{x}\in \mathcal{C}_i$}
\STATE	Update $\mathcal{D}$ using $(\bm{x}, y_{c_i})$ (Section~\ref{sec:update}).
\STATE	$\tilde{y}\leftarrow y_{c_i}$
\ENDFOR
\ENDFOR
\RETURN $Novel$

\end{algorithmic}
\end{algorithm}
%%%%%%%%%%%%%%%%%%%%%%%%%%%%%%%%%%%%%%%%%%%

\label{sec: ECD}
\subsection{Data Storage and Classifier Update (DSCU)}
\label{sec:update}
A data storage $\mathcal{D}$ is actually a storage unit consisting of $K$ buffers, where $K$ is number of classes and it stores at most $S_{\mathcal{D}}$ instances for each class. Let $\mathcal{D}_i$ denote the buffer for class $c_i$. For every $(\bm{x}, \tilde{y}=c_i)$ sent to update $\mathcal{D}$, if $\mathcal{D}_i$ is not full, $\bm{x}$ is simply added to $\mathcal{D}_i$; Otherwise, the "oldest" instance is replaced by $\bm{x}$.

The classifier $\mathit{f}$ is updated only when both of the following update conditions are satisfied. 
\begin{itemize}
\item $DetectNovel$ returns $True$.
\item Suppose $\mathit{C}$ is the set of novel classes detected by $DetectNovel$ since last update of $f$, at least one of $\mathit{C}$ should contain more than $S_{update}$ instances in $\mathcal{D}$.
\end{itemize}
If satisfied, all classes with more than $S_{update}$ instances in $\mathcal{D}$ forms a new training dataset $\mathcal{D}'$, and is then used to retrain the classifier $\mathit{f}$.

\section{Empirical Evaluation}
\label{Sec:evaluation} 
In this section, we evaluate the proposed framework on benchmark real-world datasets by comparing the performance of classification and novel class detection with other existing baseline algorithms.

\subsection{Datasets}
\label{subsec:dataset}

% \begin{table}[t]
% \centering
% \resizebox{0.7\columnwidth}{!}{%
% \begin{tabular}{|c|c|c|c|}
% \hline
% \multicolumn{1}{|c|}{\textbf{Dataset}} & \multicolumn{1}{c|}{\textbf{\# features}} & \multicolumn{1}{c|}{\textbf{\# classes}} & \multicolumn{1}{c|}{\textbf{\# instances}} \\ \hline
% % ForestCover & 54 & 7 & 150,000 \\ \hline
% FASHION-MNIST & 784 & 10 & 70,000 \\ \hline
% %Electricity & 8 & 2 & 45,311 \\ \hline
% MNIST & 784 & 10 & 70,000\\ \hline
% EMNIST & 784 & 47 & 131,600 \\ \hline
% %ForestCover & 54 & 7 & 150,000\\ \hline
% CIFAR-10 & 1024 & 10 & 60,000\\ \hline
% New York Times & 300 & 21 & 70,000\\ \hline 
% Guardian & 300 & 10 & 66,112\\ \hline
% \end{tabular}%
% }
% \caption{Description of Datasets}
% \label{tab:dataset}
% \end{table}

\begin{table}[t]
\centering
\resizebox{0.8\columnwidth}{!}{%
\begin{tabular}{|l|r|r|r|}
\hline
\multicolumn{1}{|c|}{\textbf{Dataset}} & \multicolumn{1}{c|}{\textbf{\# features}} & \multicolumn{1}{c|}{\textbf{\# classes}} & \multicolumn{1}{c|}{\textbf{\# instances}} \\ \hline
FASHION-MNIST & 784 & 10 & 70,000 \\ \cline{1-4} 
MNIST & 784 & 10 & 70,000 \\ \cline{1-4} 
EMNIST & 784 & 47 & 131,600 \\ \cline{1-4} 
CIFAR-10 & 1024 & 10 & 70,000 \\ \hline
% \multirow{2}{*}{Text} & New York Times & 300 & 21 & 70,000 \\ \cline{2-5} 
%  & Guardian & 300 & 10 & 66,112 \\ \hline
\end{tabular}%
}
\caption{Description of Datasets}
\label{tab:dataset}
\end{table}

We use four publicly available benchmark real-world image datasets including \textbf{FASHION-MNIST}~\cite{fashion-mnist}, \textbf{MNIST}~\cite{mnist}, \textbf{EMNIST}~\cite{emnist} and \textbf{CIFAR-10}~\cite{cifar10} for evaluation. The \textbf{MNIST} dataset contains $70,000$ images of handwritten digits, where each digit has been size-normalized and centered in a fixed-size image. The problem is to identify the corresponding digit for each image. \textbf{FASHION-MNIST} dataset is designed as a difficult drop-in replacement for MNIST that shares all characteristics with it, but it better represents modern computer vision (CV) tasks. Each example is a $28\times 28$ gray-scale fashion image, associated with a label from 10 classes. The \textbf{EMNIST} dataset is a set of handwritten character digits derived from the NIST Special Database 19 which contains digits, uppercase, and lowercase handwritten letters. In the experiment, we select the balanced version of EMNIST that contains $131,600$ characters with $47$ balanced classes. \textbf{CIFAR-10} is another image dataset containing $60,000$ $32\times32$ colour images in 10 classes, with $6000$ images per class. To be consistent with other datasets, we convert these images into gray-scale through OpenCV API, resulting in 1024 features. Details of these datasets are listed in Table~\ref{tab:dataset}.

An initial training set with $\lfloor n\cdot r\rfloor$ known classes is available to train the model, where $n$ is the total number of classes in the dataset and $r$ is a user-defined constant between 0 and 1 indicating the ratio of known classes in each dataset. For our experiments, we generate two streams, one with $r=0.3$ and the other with $r=0.5$. Instances of leftover classes (i.e., $n - \lfloor n\cdot r\rfloor$)  form the novel class collection. We simulate a data stream on each benchmark dataset including instances of both the known classes and new classes in the novel class collection. Note that those new classes appear in different periods in this simulated data stream with a uniform distribution. %Figure~\ref{fig:mnist_0.3_stream} illustrates the class distribution of simulated streams with $r=0.3$ on \textbf{MNIST} dataset as an example.

\subsection{Baselines}
\label{subsec:baselines}
To examine the quality of metrics learned in \sysname{}, we compare the convolutional open-world classifier learned in \sysname{} with several state-of-the-art metric learning algorithms. (1) \textbf{LMNN} (linear, EMF)~\cite{weinberger2009}: A Mahalanobis distance metric for kNN classification from labeled examples, trained with the goal that the k-nearest neighbors always belong to the same class while examples from different classes are separated by a large margin; (2) \textbf{HDML} (non-linear, EMF)~\cite{0002FS12}: A framework applicable to a broad families of mappings from high-dimensional data to binary codes that preserve semantic similarity, using a flexible form of triplet ranking loss. The mapping is represented by a well-designed hash function. In this experiment, we select the most complex hash function, i.e., multilayer neural network, proposed by the author for a fair comparison. (3) \textbf{GB-LMNN} (non-linear, EMF)~\cite{kedemgradient}: An expansion of LMNN that substitutes the linear feature mapping with non-linear Gradient Boosting Trees (GBRT). (4) \textbf{SKLR} (non-linear, IMF)~\cite{AmidGU16}: An implicit metric which learns a kernel matrix using the log-determinant divergence subject to a set of relative-distance constraints. It is useful in settings where providing similar and dissimilar constraints is difficult. 

Besides, we also compare \sysname{} with competing state-of-the-art stream classifiers. (1) \textbf{ECSMiner} (fully supervised)~\cite{masud2011}: an ensemble framework to detect novel classes using K-Means clustering, with a KNN-based classifier to make predictions; (2) \textbf{ECHO-D} (semi-supervised)~\cite{haque2016}: an improved framework based on ECSMiner that maintains an ensemble of clustering-based classifier models. Each model is trained on different dynamically-determined partially-labeled chunks of data. It detects novel classes via the same algorithm as ECSMiner but classifies instances in a different way; (3) \textbf{SENC-MaS} (semi-supervised)~\cite{mu2017streaming}: a framework that maintains two low-dimensional sketches of stream data (global and local sketch) to detect novel classes and make predictions.

\subsection{Experiment Setup}
We have implemented \sysname{} using \textit{Python} 3.6.2, and the convolutional open-world classifier using the \textit{Pytorch} 0.4.0 library. All baseline methods were based on code released by corresponding authors, except SENC-MaS. Due to unavailability of a fully functional code of SENC-MaS, we use our own implementation based on the author's description~\cite{mu2017streaming}. Hyper-parameters of these baseline approaches were set based on values reported by the authors and fine-tuned via cross-validation either on the validation dataset (metric comparison) or the initialization dataset (stream classification). In \sysname{}, we set $n=200$, $S_{\mathcal{B}}=1000$, $S_\mathcal{D}=200$, $S_{update}=100$, $\mathcal{T}_{\mathcal{D}}=0.99$, $\gamma=1.0$, $S_{\emph{mini}}=64$ and $n_{e}=10$ as default. The initial training dataset size is 1000 per class. In addition, we set the kernel size $\mathcal{K}=5$ and the stride $\mathcal{S}=1$ for convolutional layer and $\mathcal{K}=2$ and $\mathcal{S}=2$ for max-pooling layer in Conv-OWC.
% Note, in the experiment, we refer \sysname{} with $r=0.3$ as \sysname{}(0.3) and similarly \sysname{} with $r=0.5$ as \sysname{}(0.5).

\subsection{Evaluation Metrics}
\subsubsection{Stream Classification}
Let $FN$ be the total novel class instances misclassified as existing class, $FP$ be the total existing class instances misclassified as novel class, $N_c$ be the total novel class instances in the stream, and $N$ be the total number of instances in the stream. We use the following metrics to evaluate our approach and compare it with baseline methods. 
(a) \emph{Accuracy\%}: $\frac{A_{\mathrm{new}}+A_{\mathrm{known}}}{m}$,  where $A_{\mathrm{new}}$ is total number of novel class instances classified correctly, $A_{\mathrm{known}}$ is the number of known class instances identified correctly, and $m$ is the number of instances in the stream. 
(b) \emph{\% of labels}: \% of true labels requested by the framework for classifier training and update. 
(c) $M_{\mathrm{new}}$: \% of novel class instances misclassified as existing class, i.e. $\frac{FN*100}{N_c}$. 
(d) $F_{\mathrm{new}}$: \% of existing class instances misclassified as novel class, i.e. $\frac{FP*100}{N-N_c}$. 
Finally, 
% 5) Acc/Label: accuracy-label ratio. It gives the measure of amount of accuracy provided by an algorithm, given a unit amount of labels. Given two algorithms that achieve similar accuracy, the one with higher Acc/Label performs better.
(e) \emph{ratio}: $\frac{\emph{Accuracy\% of }M}{\emph{Accuracy\% of }M_{\mathrm{best}}}$, where $M$ denotes a method in \{ECSMiner, SENC-MaS, ECHO, \sysname{}\} and $M_\mathrm{best}$ is the method with the best \emph{Accuracy\%} among them.

\subsubsection{Metric Learning} 

Let $N_c$ be the total test instances belonging to class $c$, $T_c$ be the total test instances of class $c$ that are correctly predicted, and $\mathcal{C}$ be the set of all classes in the test dataset. We measure the following evaluation metric. 
% We use the following metrics to evaluate our metric learned in \sysname{} and compare it with baseline methods. 
(a) \emph{Accuracy\%}: $\frac{\sum\limits_{c\in \mathcal{C}} T_c}{\sum\limits_{c\in \mathcal{C}} N_c}$. 
(b) \emph{ratio}: $\frac{\emph{Accuracy\% of }M}{\emph{Accuracy\% of }M_{\mathrm{best}}}$, 
where $M$ denotes a method in \{LMNN, HDML, GB-LMNN, SKLR, \sysname{}\} and $M_\mathrm{best}$ is the method with the best \emph{Accuracy\%} among them.

\subsection{Results}

\begin{table*}[t]
\centering
\resizebox{\textwidth}{!}{%
\begin{tabular}{|c|r|r|r|r|r|r|r|r|r|r|r|r|}

\hline
\multirow{2}{*}{\textbf{Methods}} & \multicolumn{3}{c|}{\textbf{CIFAR-10}} & \multicolumn{3}{c|}{\textbf{MNIST}} & \multicolumn{3}{c|}{\textbf{FASHION-MNIST}} & \multicolumn{3}{c|}{\textbf{EMNIST}}\\
\cline{2-13} & Accuracy (\%) & \% of labels & \emph{ratio} & Accuracy (\%) & \% of labels & \emph{ratio} & Accuracy (\%) & \% of labels & \emph{ratio} & Accuracy (\%) & \% of labels  & \emph{ratio} \\ \hline

\textbf{ECSMiner} & 28.30$\pm$0.38 & 100.00$\pm$0.00 & 0.53 & 93.26$\pm$0.26 & 100.00$\pm$0.00 & 0.99 & 76.50$\pm$0.61 & 100.00$\pm$0.00 & 0.82 & 61.44$\pm$0.51 & 100.00$\pm$0.00 & 0.72 \\ \hline
\textbf{SENC-MaS} & 43.93$\pm$1.24 & \textbf{35.30$\pm$1.28} & 0.82 & 54.78$\pm$0.38 & 41.79$\pm$0.63 & 0.58 & 46.62$\pm$0.32 & 37.92$\pm$1.83 & 0.50 & 38.91$\pm$1.01 & \textbf{35.75$\pm$1.01} & 0.45 \\ \hline
\textbf{ECHO-D} & 27.68$\pm$0.40 & 45.34$\pm$1.01 & 0.52 & 92.64$\pm$0.26 & 47.22$\pm$1.21 & 0.98 & 68.60$\pm$1.33 & 45.11$\pm$2.09 & 0.74 & 47.06$\pm$1.99 & 43.14$\pm$1.72 & 0.55 \\ \hline
\textbf{\sysname{}} & \textbf{53.31 $\pm$ 0.63} & 39.91$\pm$0.79 & \textbf{1.00} & \textbf{94.27$\pm$0.48} &  \textbf{17.41$\pm$0.18} & \textbf{1.00} & \textbf{93.13$\pm$0.10} & \textbf{34.39$\pm$1.17} & \textbf{1.00} & \textbf{85.77$\pm$0.03} & 40.29$\pm$0.04 & \textbf{1.00} \\ \hline

\end{tabular}%
}
\caption{\textbf{Comparison of classification performance on competing methods over data streams with $r=0.3$.}}
\label{tab:result_summary_stream_0.3}
\end{table*}

\begin{table*}[t]
\centering
\resizebox{0.8\textwidth}{!}{%
\begin{tabular}{|c|r|r|r|r|r|r|r|r|}

\hline
\multirow{2}{*}{\textbf{Methods}} & \multicolumn{2}{c|}{\textbf{MNIST}} & \multicolumn{2}{c|}{\textbf{FASHION-MNIST}} & \multicolumn{2}{c|}{\textbf{EMNIST}} & \multicolumn{2}{c|}{\textbf{CIFAR-10}}\\
\cline{2-9} & $M_{\mathrm{new}}$ & $F_{\mathrm{new}}$  & $M_{\mathrm{new}}$ & $F_{\mathrm{new}}$  & $M_{\mathrm{new}}$ & $F_{\mathrm{new}}$  & $M_{\mathrm{new}}$ & $F_{\mathrm{new}}$  \\ \hline
\textbf{ECSMiner} & 66,57$\pm$5.04 & 1.10$\pm$0.01 & 100.00$\pm$0.00 & - & 100.00$\pm$0.00 & - & 100.00$\pm$0.00 & -    \\ \hline
% \textbf{SENC-MaS} & 87.673$\pm$0.66 & 83.797$\pm$2.10 & 83.823$\pm$0.75 & 63.980$\pm$2.81 & 87.333$\pm$1.41 & 81.797$\pm$1.10 & 91.892$\pm$0.74 &   92.840$\pm$0.89 \\ \hline
\textbf{SENC-MaS} & 97.76$\pm$0.10 & 5.36$\pm$0.13 & 93.76$\pm$0.34 & 20.74$\pm$0.46 & 98.22$\pm$0.01 & 6.80$\pm$0.06 & 97.06$\pm$0.22 & 1.39$\pm$0.12 \\ \hline
\textbf{ECHO-D} & 61.29$\pm$3.64 & 1.21$\pm$0.01 & 100.00$\pm$0.00 & - & 100.00$\pm$0.00 & - & 100.00$\pm$0.00 & -   \\ \hline
\textbf{\sysname{}} & \textbf{30.41$\pm$3.15} & \textbf{0.10$\pm$0.01} & \textbf{52.59$\pm$3.52} & \textbf{0.11$\pm$0.02} & \textbf{21.06$\pm$0.35} & \textbf{0.39$\pm$0.01} & \textbf{54.68$\pm$0.53} & \textbf{0.60$\pm$0.08} \\ \hline

\end{tabular}%
}
\caption{\textbf{Novel class detection performance over data streams with $r=0.3$. Here - denotes failure of novel class detection.}}
\label{tab:result_summary_novel}
\end{table*}

%%%%%%%%%%%%%%%%%%%%%%%%% Metric Learning %%%%%%%%%%%%%%%%
\begin{table*}[t]
\centering
\resizebox{0.65\textwidth}{!}{%
\begin{tabular}{|c|r|r|r|r|r|r|r|r|}
\hline
\multicolumn{9}{|c|}{\textbf{train:validation:test (4:2:4) - S1}} \\ \hline
\multirow{2}{*}{\textbf{Methods}} & \multicolumn{2}{c|}{\textbf{MNIST}} & \multicolumn{2}{c|}{\textbf{FASHION-MNIST}} & \multicolumn{2}{c|}{\textbf{EMNIST}} & \multicolumn{2}{c|}{\textbf{CIFAR-10}} \\ \cline{2-9} 
 & \multicolumn{1}{c|}{\emph{Accuracy\%}} & \multicolumn{1}{c|}{\emph{ratio}} & \multicolumn{1}{c|}{\emph{Accuracy\%}} & \multicolumn{1}{c|}{\emph{ratio}} & \multicolumn{1}{c|}{\emph{Accuracy\%}} & \multicolumn{1}{c|}{\emph{ratio}} & \multicolumn{1}{c|}{\emph{Accuracy\%}} & \multicolumn{1}{c|}{\emph{ratio}} \\ \hline
\textbf{LMNN} & 96.95$\pm$0.06 & 0.99 & 84.24$\pm$0.19 & 0.95 & 73.02$\pm$0.26 & 0.95 & 25.48$\pm$0.19 & 0.59 \\ \hline
\textbf{HDML} & 94.75$\pm$0.19 & 0.97 & 77.23$\pm$0.21 & 0.88 & 61.95$\pm$0.15 & 0.81 & 26.83$\pm$0.25 & 0.62 \\ \hline
\textbf{GB-LMNN} & 97.15$\pm$0.10 & 0.99 & 85.37$\pm$0.14 & 0.97 & 71.79$\pm$0.15 & 0.94 & 25.87$\pm$0.16 & 0.60 \\ \hline
\textbf{SKLR} & 95.65$\pm$0.06 & 0.98 & 84.47$\pm$0.09 & 0.96 & 72.48$\pm$0.15 & 0.95 & 31.98$\pm$0.11 & 0.74 \\ \hline
\textbf{\sysname{}} & \textbf{97.36$\pm$0.39} & \textbf{1.00} & \textbf{88.21$\pm$0.65} & \textbf{1.00} & \textbf{76.71$\pm$1.05} & \textbf{1.00} & \textbf{43.18$\pm$1.43} & \textbf{1.00} \\ \hline 
\hline
\multicolumn{9}{|c|}{\textbf{train:validation:test (1:1:8) - S2}} \\ \hline
\multirow{2}{*}{\textbf{Methods}} & \multicolumn{2}{c|}{\textbf{MNIST}} & \multicolumn{2}{c|}{\textbf{FASHION-MNIST}} & \multicolumn{2}{c|}{\textbf{EMNIST}} & \multicolumn{2}{c|}{\textbf{CIFAR-10}} \\ \cline{2-9} 
 & \multicolumn{1}{c|}{\emph{Accuracy\%}} & \multicolumn{1}{c|}{\emph{ratio}} & \multicolumn{1}{c|}{\emph{Accuracy\%}} & \multicolumn{1}{c|}{\emph{ratio}} & \multicolumn{1}{c|}{\emph{Accuracy\%}} & \multicolumn{1}{c|}{\emph{ratio}} & \multicolumn{1}{c|}{\emph{Accuracy\%}} & \multicolumn{1}{c|}{\emph{ratio}} \\ \hline
\multicolumn{1}{|c|}{\textbf{LMNN}} & 95.17$\pm$0.12 & 0.99 & 82.24$\pm$0.21 & 0.96  & 63.38$\pm$0.68 & 0.86 & 21.14$\pm$0.27 & 0.60 \\ \hline
\multicolumn{1}{|c|}{\textbf{HDML}} & 91.34$\pm$0.20 & 0.95 & 72.94$\pm$0.20 & 0.85 & 58.08$\pm$0.17 & 0.79 & 23.50$\pm$0.23 & 0.67 \\ \hline
\multicolumn{1}{|c|}{\textbf{GB-LMNN}} & 94.21$\pm$0.08 & 0.98 & 83.01$\pm$0.14 & 0.96 & 64.05$\pm$0.22 & 0.87 & 23.31$\pm$0.20 & 0.66 \\ \hline
\multicolumn{1}{|c|}{\textbf{SKLR}} & 93.33$\pm$0.21 & 0.97 & 81.63$\pm$0.30 & 0.95 & 63.68$\pm$0.34 & 0.86 & 29.56$\pm$0.55 & 0.84 \\ \hline
\multicolumn{1}{|c|}{\textbf{\sysname{}}} & \textbf{95.96$\pm$0.98} & \textbf{1.00} & \textbf{86.05$\pm$0.67} & \textbf{1.00} & \textbf{73.67$\pm$0.78} & \textbf{1.00} & \textbf{35.25$\pm$1.10} & \textbf{1.00} \\ \hline
\end{tabular}%
}
\caption{\textbf{Comparison of classification performance on competing metric learning algorithms.}}
\label{tab:result_summary_metric}
\end{table*}

%%%%%%%%%%%%%%%%%%% Effect of Convolutional Layers %%%%%%%%%%%%%%%%%%
\begin{table*}[t]
\centering
\begin{tabular}{|c|r|r|r|c|}
\hline
\textbf{Methods} & \emph{Accuracy\%} & $M_{\mathrm{new}}$ & $F_{\mathrm{new}}$ & Classifier Training Time (min)\\ \hline
\textbf{\sysname{}-0} & 88.97$\pm$0.46 & 56.40$\pm$2.41 & 0.16$\pm$0.03 & \textbf{1.89$\pm$0.08} \\ \hline
\textbf{\sysname{}-1} &  93.13$\pm$0.10 & 52.59$\pm$3.52 & \textbf{0.11$\pm$0.02} & 4.89$\pm$0.44 \\ \hline
\textbf{\sysname{}-2} & \textbf{93.13$\pm$0.08} & \textbf{52.51$\pm$1.64} & 0.22$\pm$0.03 & 8.06$\pm$0.08 \\ \hline
\end{tabular}
\caption{Effect of convolutional layers on classification and novel class detection performances over the FASHION-MNIST data stream with $r=0.3$. Numbers in the names indicate the number of convolutional layers used in \sysname{}.}
\label{tab:conv_effect}
\end{table*}
%%%%%%%%%%%%%%%%%%% News Stream %%%%%%%%%%%%%%%%%%%%%%%%%%
\begin{figure}%
    \centering
    {{\includegraphics[width=0.95\columnwidth]{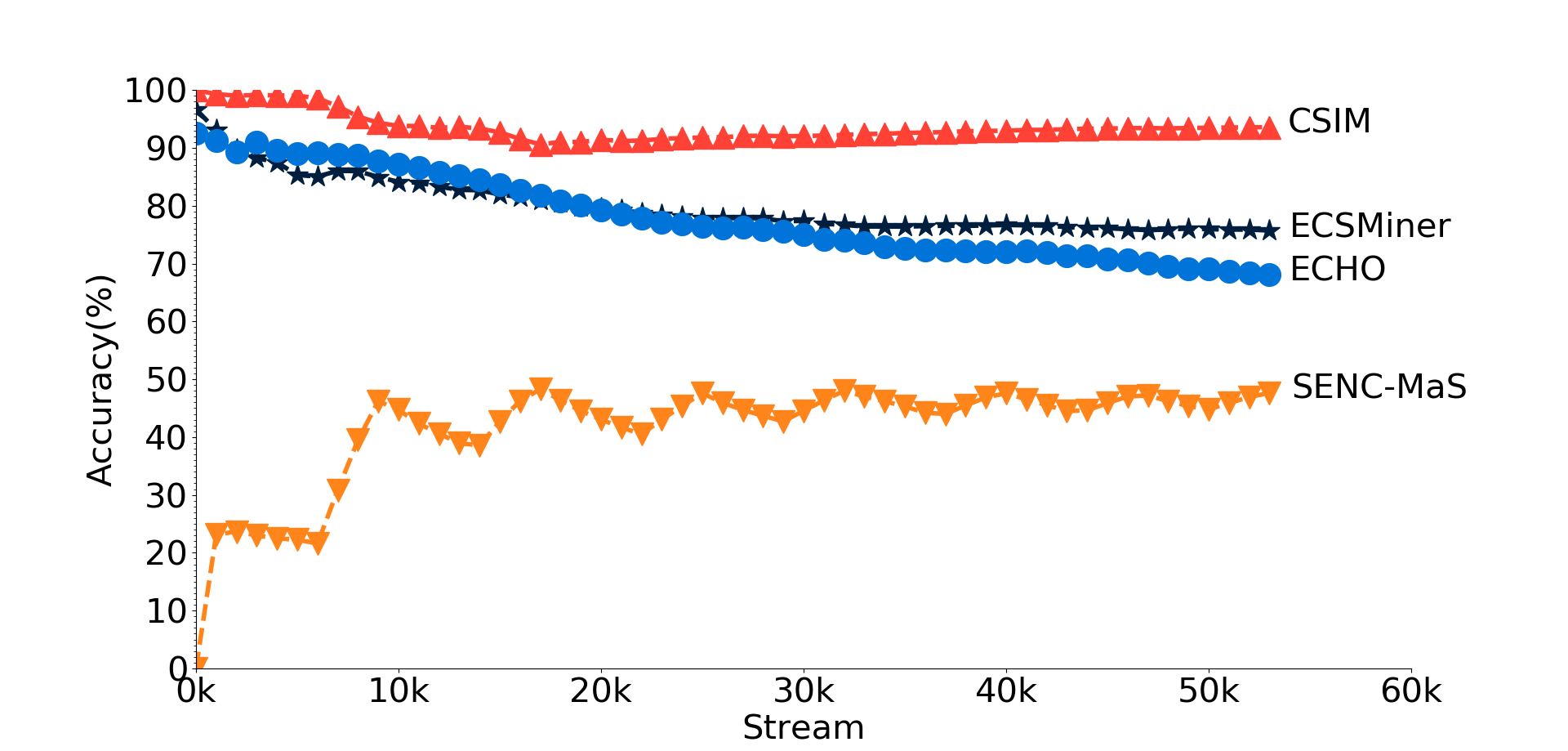} }}%
    \qquad
    \caption{Accuracy result over the FASHION-MNIST data stream with $r=0.3$.}%
    \label{fig:fashionmnist_stream}%
\end{figure}

\begin{figure}[t]
\centering
\includegraphics[width=1.0\columnwidth]{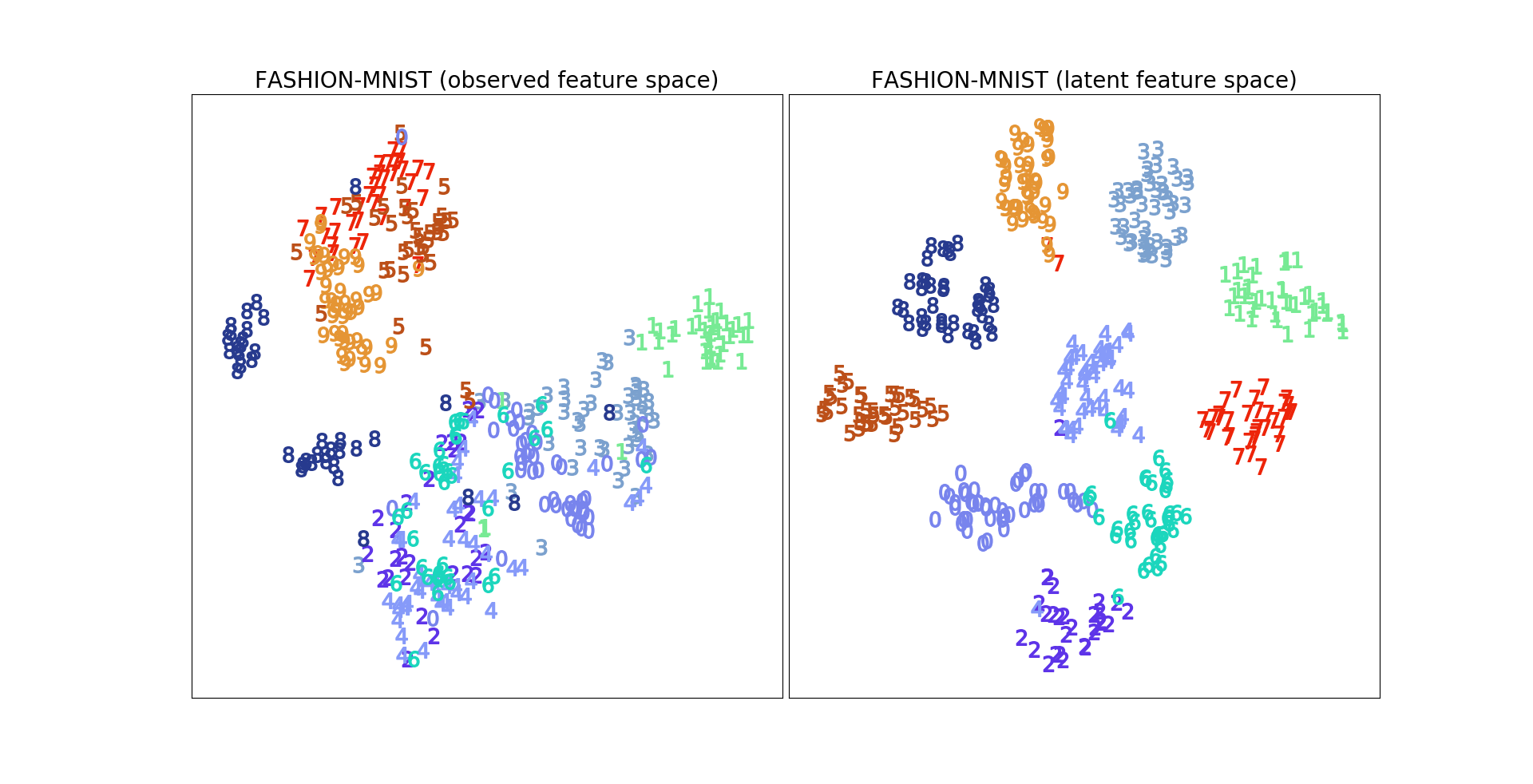}
\caption{TSNE graph of embeddings in observed feature space (left) and transformed latent feature space (right) on FASHION-MNIST dataset.}
\label{fig:metric_transform}
\end{figure}
% %%%%%%%%%%%%%%%% Statistics %%%%%%%%%%%%%%%%%%%%%%%%
% \begin{table}[t]
% \centering
% \scalebox{0.9} {%
% \begin{tabular}{|c|c|c|c|c|}
% \hline
% \textbf{Algorithm} & \textbf{MNIST} & \textbf{FASHION-MNIST} & \textbf{EMNIST} & \textbf{CIFAR-10} \\ \hline
% % \textbf{AHT} & 1.02$e$-4 & 9.95$e$-11 & 3.36$e$-4 &  1.05$e$-8  \\ \hline
% \textbf{ECSMiner} &  &  &  &    \\ \hline
% \textbf{SENC-MaS} &  &  &  &    \\ \hline
% \textbf{SAND-D} &  &  &  &   \\ \hline
% \end{tabular}
% }
% \caption{\textbf{P-value test on accuracy between \sysname{} and baselines}.}
% \label{tab: statistics}
% \end{table}

%%%%%%%%%%%%%%%%%% Sensitivity %%%%%%%%%%%%%%%%%%%%
\begin{figure}[t]
\centering
\includegraphics[width=\columnwidth]{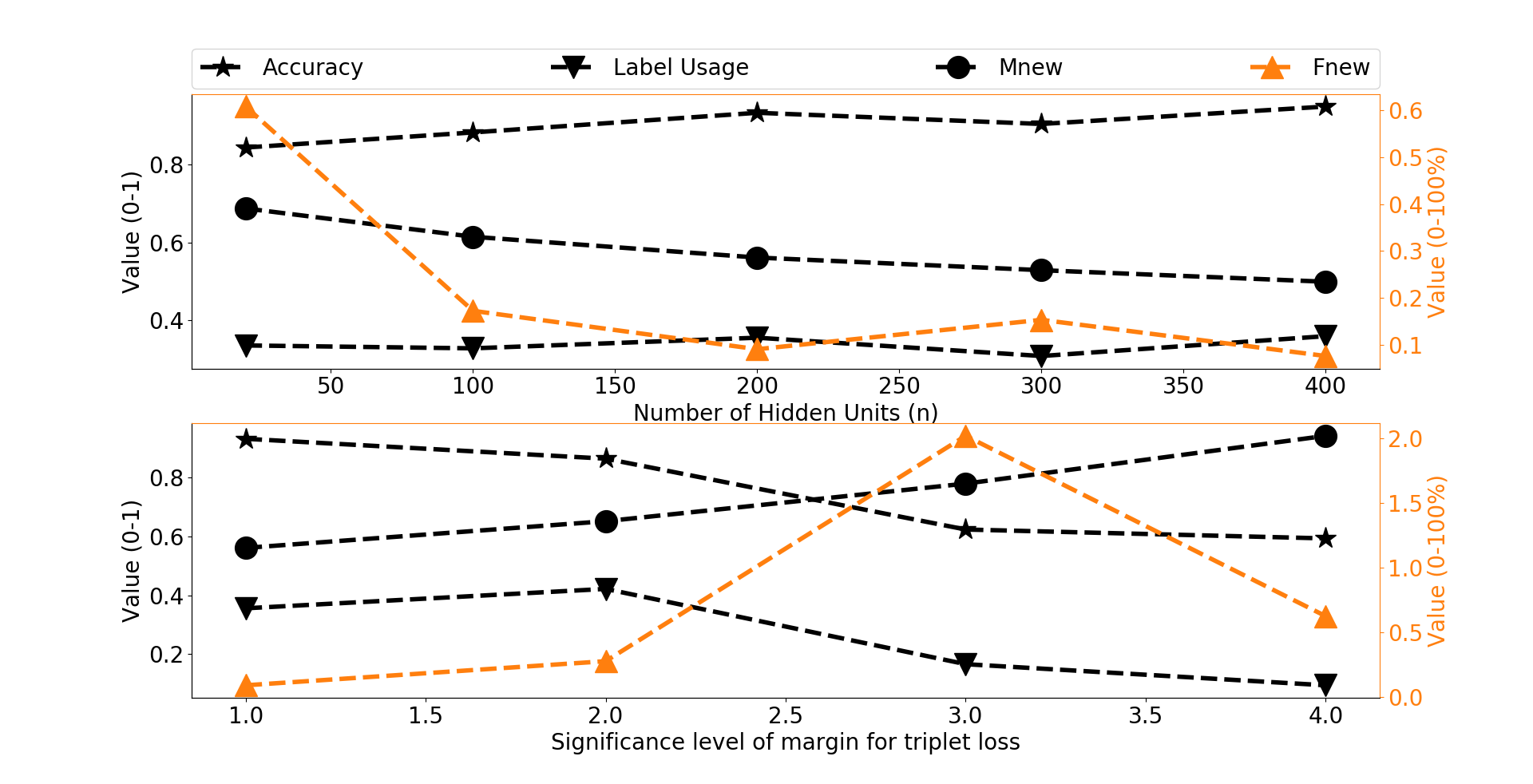}
\caption{Parameter sensitivity (n and $\gamma$) of \sysname{} for FASHION-MNIST dataset as an example.}
\label{fig:sensitivity}
\end{figure}

% \subsubsection{Metric Learning} In the experiment, we first randomly shuffle each benchmark dataset and then divide it into training, validation and test sets. We select the split ratio to be $4:2:4$ and $1:1:8$ to study the generalization performance of learned metrics on unseen data when limited training data is available, which is a common case on streams. This process is repeated for $10$ times to avoid any statistical fluctuation. Both mean and standard deviation of performance are reported in Table~\ref{tab:result_summary_metric}

\subsubsection{Stream Classification}
% The experiments on each benchmark dataset are repeated for $10$ times with different simulated streams and both the mean and the standard deviation of the performance are reported.
We conduct $10$ independent experiments with different simulated streams for both $r=0.3$ and $r=0.5$ on each real-world benchmark dataset. However, we only report the mean and standard deviation of performance on streams with $r=0.3$ due to lack of space, though we observed similar result on $r=0.5$. Table~\ref{tab:result_summary_stream_0.3} lists the results on data streams with $r=0.3$. As mentioned in Section~\ref{subsec:dataset}, the $r$ value indicates the number of classes known in the warm-up phase. For example, with $r=0.3$, only 3 classes are known in the initial training data on MNIST. We observe that SENC-MaS performs poorly on most real-world datasets due to its linear classifier. ECSMiner performs better than ECHO-D because the former is fully-supervised and the latter is semi-supervised.
However, in both cases of $r$, \sysname{} outperforms all the baseline approaches by providing significantly better accuracy while requesting fewer or similar amount of true labels. For example, on EMNIST dataset ($r=0.3$, $47$ classes), \sysname{} provides an accuracy of $85.36\%$, which is $23.92\%$ higher than that provided by the best baseline ECSMiner and reduces the number of ground truth labels requested by $60.07\%$.
We observe similar results for EMNIST stream with $r=0.5$.
\sysname{} is much better than all baselines mainly because of the intrinsic similarity metric learned via multi-task learning which improves the performance of both classification and novel class detection. Moreover, the convolutional layer in Conv-OWC aids in detecting edges in images which forms the conceptual representation that helps to improve the quality of learned intrinsic similarity metric.
% We will study the quality of learned EMF and compare it with other metric learning baselines in Section~\ref{subsubsec:ML}.

% \mbox{}
\subsubsection{Novel Class Detection}
Table~\ref{tab:result_summary_novel} compares novel class detection performance of \sysname{} with all baseline methods on each dataset. We observe that both ECSMiner and ECHO fails to detect any novel class on most real-world datasets. On the other hand, SENC-MaS could detect some novel class instances with poor precision while missing most of such instances.  It is because all these approaches rely on the strong global cohesion and separation assumption which is invalid for many complex real-world datasets we use. In contrast, \sysname{} relaxes this strong assumption and provides the lowest $M_{\mathrm{new}}$ and $F_{\mathrm{new}}$ compared to these baselines. Hence \sysname{} outperforms in not only detecting more true novel class instances but also providing a lower false alarm rate, which is desired in a novel class detection task.

\mbox{\tiny}
\subsubsection{Stability of \sysname{} over Data Streams}
Figure~\ref{fig:fashionmnist_stream} shows the classification accuracy of \sysname{} over the FASHION-MNIST data stream. As shown in the figure, \sysname{} performs better than baselines and maintains good performance with new classes continuously emerging over time. In particular, \sysname{} adapts to the occurrence of unknown classes quickly compared to SENC-MaS and produces more accurate predictions. ECHO shows an unstable performance that degrades significantly and ECSMiner results in slightly better performance compared to ECHO since it is fully-supervised. Similar results have been observed on other data streams.

\mbox{\tiny}
\subsubsection{Metric Learning}
\label{subsubsec:ML}

To study the generalization performance of learned metrics in Conv-OWC on unseen data when limited training data is available, a common case on streams, we perform experiments on the two splits (i.e., $S1$ and $S2$) over the image datasets. Here, we first randomly shuffle each benchmark dataset and then divide it into training, validation and test sets with the split ratio of $4:2:4$ and $1:1:8$. We denote these splits as $S1$ and $S2$ respectively. Here, $S2$ is more realistic for a data stream.
This process is repeated for $10$ times to avoid any statistical fluctuation. Both mean and standard deviation of performance are reported in Table~\ref{tab:result_summary_metric}. As shown in the result, for both ratios, \sysname{} outperforms all baseline approaches on all benchmark datasets by providing significantly better classification accuracy. For example, on EMNIST dataset that contains 47 classes, \sysname{} provides a higher accuracy of $73.67\%$ compared to the best baseline GB-LMNN with a margin of $9.62\%$ for the $S2$ split. The superior performance of \sysname{} demonstrates its better capability of capturing intra-class similarity and inter-class dissimilarity with a limited amount of labeled training data. Fig.~\ref{fig:metric_transform} illustrates an example of original and transformed embeddings provided by \sysname{} on FASHION-MNIST dataset. Hence, compared to other state-of-the-art baselines, our proposed metric learning approach is more suitable for stream applications.

\mbox{}
\subsubsection{Sensitivity of Parameters}
The two main parameters in \sysname{} are the number of hidden units $n$ in the Conv-OWC, and significance level $\gamma$ of margin for triplet loss. We vary these parameters to study its sensitivity to classification and novel class detection performance. Figure~\ref{fig:sensitivity} shows the result on FASHION-MNIST dataset as an example. If $n$ is relatively small, it indicates a simple network. In this case, the classification and novel class detection performance significantly drops by providing a lower accuracy and higher $M_{\mathrm{new}}$ and $F_{\mathrm{new}}$. On the other hand, a larger $n$ reduces $M_{\mathrm{new}}$ but provides little improvement on the other metrics and dramatically increases the time and space cost. Similarly, as $\gamma$ increases, \sysname{} attempts to push different classes with a margin that is too large, leading to overfitting issues. Therefore, we choose a moderate value of $n=200$ and $\gamma=1.0$ during evaluation. \looseness=-1

\mbox{}
\subsubsection{Effect of Convolutional Layers}
To study the effect of convolutional layers on both classification and novel class detection performance over data streams, we built two variants of Conv-OWC with 0 (\sysname{}-0) and 2 (\sysname{}-2) convolutional layers respectively. Table~\ref{tab:conv_effect} reports the results on FASHION-MNIST data stream as an example. A significant improvement on classification and novel class detection performance is observed from \sysname{}-1 to \sysname{}-0, which indicates the edges recognized by the convolutional layer actually reduces the difficulty of subsequent metric learning task. However, adding more convolutional layers provides little help for performance improvement but dramatically increases the training time and is hence undesired. It is mainly because of the limited amount of training data along the stream that is insufficient for a bigger network to improve the quality of its conceptual representation.
Therefore, our choice of single convolutional layer in \sysname{} is recommended.

\mbox{}
\subsubsection{Time and Space Complexity and Limitation}
\label{sec: TSC}
Overall, the execution overhead of \sysname{} mainly arises from the training and updating procedure, particularly while training the convolutional open-world classifier. Assuming that the time complexity of calculating the gradient of one example is a constant $C$, the time complexity of MBGD within a mini-batch is $O(S_{\mathrm{mini}}^3C)$. Thus the total time complexity of \sysname{} becomes $O(n_eS_{\mathrm{mini}}^2CS_{\mathcal{D}}||\mathcal{Y}'||)$, where $||\mathcal{Y}'||$ denotes the number of classes in $\mathcal{Y}'$. Clearly, the large overhead of \sysname{} mainly comes from the gradient computing in each mini-batch. 
% Fortunately, this overhead can be greatly reduced by accelerating the computation with a GPU. 
In our implementation, we use a GPU for computational acceleration.
By utilizing a GTX 1080 Ti 11GB GPU, the average training time for \sysname{} is $29.33$ seconds per epoch and hence total training time is approximately $4.89$ minutes. The space complexity of \sysname{} is $O(S_{\mathcal{B}}+S_{\mathcal{D}}||\mathcal{Y}'||+B_{\mathrm{space}})$, where $B_{\mathrm{space}}$ denotes the space complexity of the model used to represent $\phi$. 

Although \sysname{} demonstrates a good performance on many real-world stream application tasks, it has several drawbacks. First, \sysname{} relies on the quality of true labels. An error in ground truth labels reduces the quality of learned metrics and degrades the classification performance. Second, \sysname{} requires more computational resources for execution compared to other approaches due to the use of a neural network. We leave the exploration for other non-linear kernel-based approaches, that can replace the neural network, for future work.

\section{Conclusions}
In this paper, we propose a novel semi-supervised stream classification framework that utilizes a convolutional open-world classifier with an intrinsic high-quality similarity metric trained via multi-task learning. This framework addresses the challenge of novel class detection problem with better performance compared to state-of-the-art baselines. More importantly, we discard the strong global class cohesion and separation assumption in novel class detection and demonstrate a technique to detect instances from multiple new classes using the convolutional open-world classifier. Our empirical evaluation of real-world datasets and streams shows the practical benefit of \sysname{} as we compare our results with state-of-the-art stream mining systems.\looseness=-1

% We could extend the work in the following direction. As discussed in Section~\ref{sec:ML}, a single-layer neural network is applied to represent the EMF. However, any non-linear function can be used here. We will explore the possibility of choosing other functions, like gradient-boosted trees. In addition, \sysname{} only addresses the problem of novel class detection. Thus, a study of incorporating concept drift module in \sysname{} is necessary, since concept drift is common in real-world streams.
\label{Sec:conclusion}

\section{Acknowledgments}
We thank the reviewers for their insightful comments. This material is based upon work supported by NSF award number DMS-1737978, AFOSR award number FA9550-14-1-0173, NSA and IBM faculty award (Research).

\bibliographystyle{ACM-Reference-Format}
\bibliography{reference.bib}

\end{document}